\documentclass[lettersize,journal,twoside]{IEEEtran}

\usepackage{amsmath,amsfonts}
\usepackage{algorithmic}
\usepackage{algorithm}
\usepackage{array}
\usepackage[caption=false,font=normalsize,labelfont=sf,textfont=sf]{subfig}
\usepackage{textcomp}
\usepackage{stfloats}
\usepackage{url}
\usepackage{verbatim}
\usepackage{graphicx}
\usepackage{cite}
\usepackage{booktabs}
\usepackage{xcolor}

\usepackage[hidelinks]{hyperref}
\usepackage{xurl}

\begin{document}

\title{Understanding the Effect of Visual Attention in CNNs: A Decision-Space Alignment Perspective}

\author{Niladri B. Puhan, and K. Kamala Priya%
\thanks{Niladri B. Puhan and K. Kamala Priya are with the School of Electrical and Computer Sciences, Indian Institute of Technology Bhubaneswar, 752050, India
(e-mail: nbpuhan@iitbbs.ac.in; a25ec09001@iitbbs.ac.in).}}

\maketitle

\begin{abstract}
Visual Attention enhances convolutional neural networks (CNNs) by modulating intermediate features, yet their effects on downstream decisions remain less understood. We present a decision-space framework that models multiplicative attention as an intermediate feature perturbation and derives a first-order prediction of its effect on the target-versus-competitor classification margin. We further introduce the Decision Alignment Score (DAS) to quantify the directional alignment between the perturbation and the local margin gradient. We evaluate the framework on CIFAR-100 using SimAM and CBAM across five popular CNN backbones: ResNet-18, ResNet-20, ResNet-50, ResNet-56, and MobileNetV2. The analysis examines prediction agreement, DAS--margin correlations, and downstream activation-state changes. Across most final-attention configurations, predicted and actual margin changes show good agreement, while DAS exhibits positive correlations with actual margin changes. Single-attention experiments further show that prediction agreement and DAS–margin correlations depend on attention placement, with stronger associations at later stages.

\end{abstract}


\section{Introduction}
\label{sec:introduction}

Many attention mechanisms enhance convolutional neural networks (CNNs) by adaptively modulating intermediate feature representations. Popular methods include channel-wise attention in squeeze-and-excitation networks~\cite{hu2018squeeze}, joint channel and spatial attention in CBAM~\cite{woo2018cbam}, and parameter-free neuron-wise attention in SimAM~\cite{yang2021simam}. Despite their effectiveness in visual recognition, only classification accuracy and feature visualization provide limited insights into how attention-induced modifications influence downstream decisions.

Feature modulation does not necessarily imply a change in the classification decision. Attention weights indicate which feature responses are emphasized or suppressed, but not how these modifications affect the classification decision. Their impact depends on both the feature perturbation and its propagation through subsequent network layers. Thus, emphasizing a feature does not necessarily improve the decision, motivating an analysis that directly relates attention-induced modifications to downstream decision behavior.

We formulate multiplicative attention as a perturbation of an intermediate feature representation and quantify its effect through the change in the target-versus-competitor classification margin. For downstream CNNs with affine operations and piecewise-linear activations, the mapping is locally affine within a fixed activation region~\cite{montufar2014regions}. This enables a first-order approximation of the margin change using its gradient with respect to intermediate features, allowing us to compare the predicted response with the actual effect of attention.

The Decision Alignment Score (DAS) is introduced to quantify the directional alignment between the attention-induced perturbation and the local decision direction. While the first-order predicted margin change incorporates both vector magnitudes and their alignment, DAS isolates their directional relationship, indicating whether the perturbation acts along or against the direction of increasing classification margin. Together, these measures characterize the predicted margin response and its underlying directional alignment. Comparing predicted and actual margin changes further assesses the accuracy of the local approximation across attention configurations.

We evaluate the framework using SimAM and CBAM across five CNN backbones (ResNet-18, ResNet-20, ResNet-50, ResNet-56, and MobileNetV2) on the CIFAR-100 dataset. Two configurations are considered: (i) ALL, where attention is applied in every residual or inverted-residual block, and the final module is analyzed while preceding attention operations remain active; and (ii) Single-attention modules, with one attention module independently placed at each of four stages in a CNN. We examine prediction agreement, the association between DAS and actual margin changes, and downstream activation-state changes across these diverse configurations.
Overall, this work provides a decision-space perspective for examining how attention mechanisms affect CNN classification behavior. The framework goes beyond identifying which features are modulated to investigate how these modifications influence downstream decisions.

The main contributions of this work are summarized as follows:
\begin{itemize}
\item We formulate multiplicative attention as an intermediate feature perturbation and characterize its downstream effect through the actual change in the target-versus-competitor classification margin.

\item We derive a first-order prediction of attention-induced margin change and introduce the Decision Alignment Score (DAS) to quantify the directional alignment between the feature perturbation and the local decision direction.

\item We systematically evaluate the proposed framework using SimAM and CBAM across five CNN backbones on CIFAR-100, investigating prediction agreement, DAS association, and downstream activation-state changes under ALL-last and  single-attention module configurations.
\end{itemize}

\section{Related Work}
\label{sec:related_work}

\subsection{Attention Mechanisms in CNNs}
\label{sec:attention_related}

Squeeze-and-Excitation (SE) networks introduced channel-wise recalibration through learned channel dependencies~\cite{hu2018squeeze}, while CBAM extended this formulation by sequentially incorporating channel and spatial attention~\cite{woo2018cbam}. ECA-Net further improved channel attention through local cross-channel interactions without dimensionality reduction~\cite{wang2020eca}. SimAM provides parameter-free, neuron-wise attention through an energy-based formulation~\cite{yang2021simam}. These approaches demonstrate how selective feature modulation can improve CNN representations and recognition performance. A comprehensive survey by Guo et al.~\cite{guo2022attention} reviews attention mechanisms in computer vision, covering channel, spatial, temporal, and branch attention.

Puhan and coauthors have investigated attention-based feature learning across diverse applications. Their studies include interleaved and multi-deformation-aware attention for structural defect classification~\cite{bhattacharya2021interleaved,bhattacharya2021multideformation}, composite attention and spatial--channel interactions for discriminative feature learning~\cite{bhattacharya2022standalone,bhattacharya2022kernelized}, and perturbed, high-boost, and transform-domain attention mechanisms for cross-spectral periocular recognition~\cite{behera2022perturbed,behera2022highboost,behera2024dct}. Perceptually-guided attention has also been investigated for medical image segmentation~\cite{bhargav2024jnd}. These studies demonstrate the application of attention formulations for task-relevant feature learning.

Despite differing attention-creation strategies, these methods share a common principle: multiplicative modulation of intermediate features. However, recognition performance and attention map visualization do not directly reveal how such modifications affect downstream decisions. We address this gap by examining the decision-margin response to attention-induced feature perturbations.

\subsection{Gradient Alignment and Feature Perturbation Analysis}
\label{subsec:related-decision-space}

Gradient directions and feature perturbations have been explored to understand model sensitivity and decision behavior. Lanfredi et al.~\cite{lanfredi2023quantifying} investigated gradient alignment in adversarial training, examining the directional properties of model gradients in relation to robustness. Yang et al.~\cite{yang2026attention} studied attention-inverted feature perturbations for semi-supervised medical image segmentation. These studies examine how gradient directions and feature perturbations relate to network behavior.

\subsection{CNN Interpretation and Feature-Space Analysis}
\label{sec:interpretability_related}

CNN interpretability has been extensively studied through feature attribution, visualization, and perturbation-based methods. Gradient-based approaches characterize the sensitivity of network predictions to input or intermediate representations. Simonyan et al.~\cite{simonyan2014deep} introduced class-specific saliency maps based on input gradients, while Sundararajan et al.~\cite{sundararajan2017axiomatic} proposed Integrated Gradients using attribution axioms. Grad-CAM uses class-specific gradients to generate localization maps highlighting image regions relevant to a target prediction~\cite{selvaraju2017gradcam}.

Perturbation-based approaches examine prediction responses to controlled modifications of input features. Fong and Vedaldi~\cite{fong2017meaningful} introduced meaningful perturbations to identify image regions responsible for classifier decisions. Separately, the piecewise-linear structure of ReLU networks has been studied theoretically~\cite{arora2018understanding, montufar2014regions}, providing a basis for analyzing local network responses to feature perturbations.

Unlike attribution and perturbation-based methods that assess feature relevance or prediction sensitivity, our approach examines how attention-induced feature perturbations affect the target–competitor margin. By comparing actual and first-order predicted margin changes and measuring their directional alignment through DAS, we provide a complementary decision-space perspective on visual attention.

\section{Proposed Methodology}
\label{sec:methodology}

\begin{figure*}[]
\centering
\includegraphics[
    width=0.9\textwidth,
    keepaspectratio
]{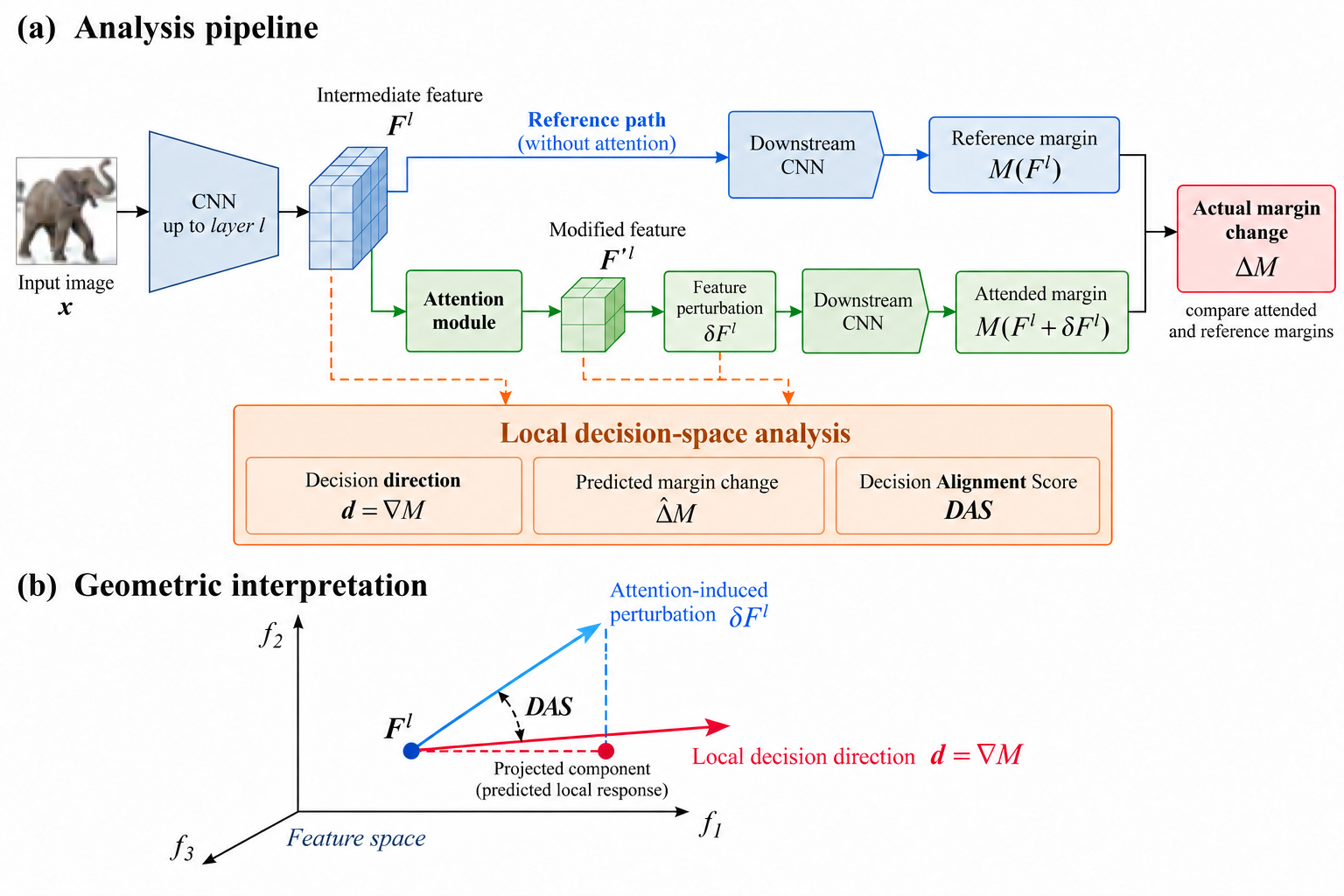}
\caption{Conceptual illustration of attention-induced feature perturbations. (a) Analysis of actual and predicted margin changes and the Decision Alignment Score (DAS). (b) Geometric interpretation of the relationship between the feature perturbation and the local decision direction.}

\label{fig:decision-space-concept}
\end{figure*}

We present a decision-space framework that models attention-induced feature modifications as intermediate perturbations and analyzes their effects on classification margins. Figure~\ref{fig:decision-space-concept} illustrates the actual margin response, its prediction, and the corresponding alignment.

\subsection{Attention-Induced Feature Perturbation}
\label{sec:attention_perturbation}

Consider an intermediate feature representation at layer
$\ell$ of a trained CNN,
\begin{equation}
F^\ell\in\mathbb{R}^{C\times H\times W},
\end{equation}
where $C$, $H$, and $W$ denote the number of channels, height, and width, respectively. For mathematical analysis, the feature tensor is vectorized as
$F^\ell\in\mathbb{R}^{N}$, where $N=CHW$.

A multiplicative attention mechanism produces attention weights $W^\ell\in\mathbb{R}^{N}$ and modifies the intermediate feature representation as
\begin{equation}
F^{\ell\prime}=W^\ell\odot F^\ell,
\label{eq:attention_operation}
\end{equation}
where $\odot$ denotes element-wise multiplication.

The modification introduced by attention can be expressed as a feature perturbation:
\begin{equation}
\delta F^\ell=F^{\ell\prime}-F^\ell.
\end{equation}
Substituting (\ref{eq:attention_operation}) gives
\begin{equation}
{
\delta F^\ell=(W^\ell-\mathbf{1})\odot F^\ell,
}
\label{eq:feature_perturbation}
\end{equation}
where $\mathbf{1}$ denotes an all-one vector.

\subsection{Classification Margin and Actual Decision Response}
\label{sec:actual_response}

Let $G_{\theta,\ell}(\cdot)$ denote the downstream mapping from the selected intermediate representation to the output logits of the trained CNN,
\begin{equation}
Z=G_{\theta,\ell}(F^\ell),
\label{eq:downstream_mapping}
\end{equation}
where $Z=[z_1,\ldots,z_K]^T$ contains the logits for
$K$ classes and $\theta$ denotes the fixed trained
network parameters.

For the target class $y$, the strongest competing class
$c$ is selected from the reference logits as
\begin{equation}
c=\arg\max_{k\neq y}z_k.
\label{eq:competitor}
\end{equation}
The class pair $(y,c)$ is kept fixed when computing the predicted and actual margin changes.

The classification margin is defined as
\begin{equation}
M(F^\ell)=z_y(F^\ell)-z_c(F^\ell).
\label{eq:decision_margin}
\end{equation}

To quantify the decision response caused by attention, we evaluate the downstream network at both the original and attention-modified feature representations. The actual margin change is defined as
\begin{equation}
{
\Delta M=
M(F^\ell+\delta F^\ell)-M(F^\ell).
}
\label{eq:actual_margin}
\end{equation}

\subsection{First-Order Prediction of Margin Change}
\label{sec:local_prediction}

We next establish the relationship between the attention-induced feature perturbation and the resulting classification margin change using the local behavior of the CNN. For CNNs composed of affine operations and piecewise-linear activations, such as ReLU and ReLU6, the downstream mapping is affine within a fixed activation region~\cite{montufar2014regions, arora2018understanding}. Hence, locally,
\begin{equation}
G_{\theta,\ell}(F^\ell)
=
A_{\theta,\ell}F^\ell+b_{\theta,\ell},
\label{eq:local_affine}
\end{equation}
where $A_{\theta,\ell}$ and $b_{\theta,\ell}$ are determined by the trained downstream network and its corresponding activation region.

If the perturbation path from $F^\ell$ to
$F^\ell+\delta F^\ell$ remains within the same
affine region, the change in the output logits is given by

\begin{equation}
\delta Z
=
A_{\theta,\ell}\delta F^\ell.
\label{eq:logit_response}
\end{equation}

For class pair $(y,c)$, define a vector $e\in\mathbb{R}^{K}$ as
\begin{equation}
e_k=
\begin{cases}
1, & k=y,\\
-1, & k=c,\\
0, & \text{otherwise}.
\end{cases}
\label{eq:class_vector}
\end{equation}

The classification margin defined in (\ref{eq:decision_margin}) can equivalently be expressed as $M(F^\ell)=e^TZ$. Accordingly, the margin change within the same affine region is
\begin{equation}
\delta M
=
e^T\delta Z
=
e^TA_{\theta,\ell}\delta F^\ell.
\label{eq:local_margin_response}
\end{equation}

We define the local decision direction as
\begin{equation}
d_{\theta,\ell}
=
A_{\theta,\ell}^{T}e.
\label{eq:decision_direction}
\end{equation}

The vector $d_{\theta,\ell}$ characterizes the local sensitivity of the target--competitor margin to changes in the intermediate feature representation. It can be computed directly as the gradient of the classification margin:
\begin{equation}
d_{\theta,\ell}
=
\nabla_{F^\ell}M.
\label{eq:gradient_direction}
\end{equation}

Consequently, the margin change within the same affine region is given by
$d_{\theta,\ell}^{T}\delta F^\ell$.
For a perturbation crossing activation-region boundaries, the downstream mapping may change along the perturbation path.

We therefore use the local margin response to define the first-order predicted margin change:
\begin{equation}
{
\widehat{\Delta M}
=
d_{\theta,\ell}^{T}\delta F^\ell.
}
\label{eq:predicted_margin}
\end{equation}
This prediction is exact when the perturbation path remains within the same affine region and otherwise provides a local approximation to the actual margin change.

Substituting the attention-induced feature perturbation from (\ref{eq:feature_perturbation}) gives
\begin{equation}
\widehat{\Delta M}
=
d_{\theta,\ell}^{T}
\left[(W^\ell-\mathbf{1})\odot F^\ell\right].
\label{eq:attention_margin_prediction}
\end{equation}

\subsection{Decision Alignment Score}
\label{sec:decision_alignment}

The predicted margin change is determined by projecting the attention-induced perturbation onto the local decision direction. This projection combines the directional relationship between the two vectors with their respective magnitudes.

To quantify their directional relationship independently of magnitude, we define the Decision Alignment Score (DAS) as
\begin{equation}
{
\Psi=
\frac{
d_{\theta,\ell}^{T}\delta F^\ell
}{
\|d_{\theta,\ell}\|_2
\|\delta F^\ell\|_2
}.
}
\label{eq:das}
\end{equation}

The score represents the cosine similarity between the attention-induced feature perturbation and the local decision direction, with
$-1\leq\Psi\leq1$.

A positive DAS indicates that the perturbation is aligned with the local direction of increasing classification margin, while a negative DAS indicates an opposing direction. A DAS close to zero indicates that the perturbation is approximately orthogonal to the local decision direction.

The relationship between DAS and the predicted margin change can be expressed as
\begin{equation}
\widehat{\Delta M}
=
\|d_{\theta,\ell}\|_2
\|\delta F^\ell\|_2
\Psi.
\label{eq:margin_decomposition}
\end{equation}

Equation (\ref{eq:margin_decomposition}) decomposes the predicted margin change into three components: gradient magnitude $\|d_{\theta,\ell}\|_2$, representing local margin sensitivity; perturbation magnitude
$\|\delta F^\ell\|_2$, quantifying the extent of feature
modification; and DAS $\Psi$, measuring directional alignment with the local decision direction.

Two attention-induced perturbations may produce similar predicted margin changes despite different DAS values, owing to variations in gradient sensitivity and perturbation magnitude. Thus, while $\widehat{\Delta M}$ captures the magnitude and direction of the predicted margin response, DAS isolates directional alignment. These quantities are mathematically related but provide complementary information.

\subsection{Computation Steps}
\label{sec:practical_computation}

For a trained CNN and a given input, the feature representation $F^\ell$ immediately before the selected attention operation is obtained. The attention-induced perturbation $\delta F^\ell$ is computed using (\ref{eq:feature_perturbation}). The local decision direction $d_{\theta,\ell}$ is computed from the gradient of the target-versus-competitor margin at the pre-attention representation, with the selected attention operation bypassed. The predicted margin change $\widehat{\Delta M}$, actual margin change
$\Delta M$, and DAS $\Psi$ are then obtained using
(\ref{eq:predicted_margin}), (\ref{eq:actual_margin}), and (\ref{eq:das}), respectively.

Prediction agreement is evaluated using Pearson and Spearman correlations, mean absolute error (MAE), root mean squared error (RMSE), and sign agreement. The relationship between DAS and actual margin change is examined using correlation analysis. All quantities are computed with the trained CNN  fixed, without introducing additional trainable parameters.

\section{Experimental Setup}
\label{sec:cifar100-results}
\begin{figure*}[]
\centering
\includegraphics[
    width=\textwidth,
    keepaspectratio
]{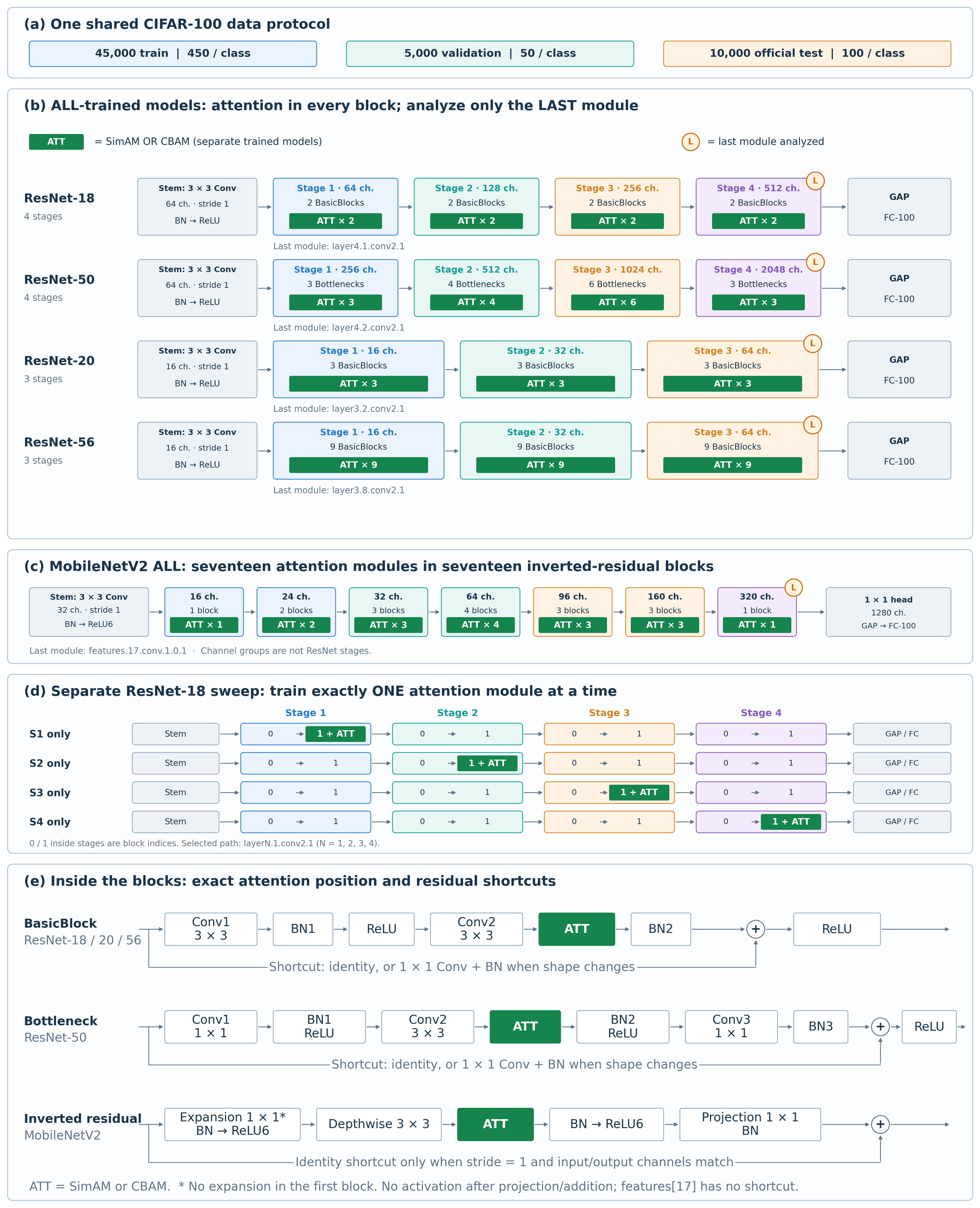}
\caption{Backbone architectures with different attention configurations. The ALL configuration inserts attention in every block, with only the final module bypassed for analysis. S1–S4 place one attention module in the final block of the corresponding ResNet-18 stage. Green indicates SimAM or CBAM.}
\label{fig:cifar-architecture}
\end{figure*}

\subsection{Dataset and Configurations}

We conduct experiments on CIFAR-100~\cite{krizhevsky2009learning}, which comprises 60,000 RGB images of size $32\times32$ across 100 classes. The official split contains 50,000 training and 10,000 test images. We further divide the training set into 45,000 training and 5,000 validation images using a fixed stratified split, with 450 and 50 images per class, respectively. The same split is used across all backbones, while the test set is used for evaluation.

We consider five CNN backbones: ResNet-18, ResNet-20, ResNet-50, ResNet-56, and MobileNetV2, using SimAM and CBAM as attention mechanisms. For each backbone--attention pair, the ALL configuration inserts attention into every residual or inverted-residual block. To investigate the effect of attention placement, we additionally train four ResNet-18 models independently for each attention method, each containing a single module in the last block of one of the four stages (S1--S4). An attention-free ResNet-18 model (B0), trained separately, serves as the baseline for recognition comparisons. Figure~ \ref{fig:cifar-architecture} illustrates the ALL and single-attention module configurations.

\begin{table}[]
\centering
\caption{Classification accuracy (\%) of the attention-free baseline (B0) and ALL-trained models across five CNN backbones on CIFAR-100.}
\label{tab:all-recognition}
\footnotesize
\setlength{\tabcolsep}{0pt}
\renewcommand{\arraystretch}{1.1}

\begin{tabular*}{\columnwidth}
{@{\extracolsep{\fill}}lrrrrrr@{}}
\toprule
& \multicolumn{2}{c}{B0}
& \multicolumn{2}{c}{SimAM ALL}
& \multicolumn{2}{c}{CBAM ALL}\\
\cmidrule(lr){2-3}
\cmidrule(lr){4-5}
\cmidrule(lr){6-7}
Backbone
& Top-1 & Top-5
& Top-1 & Top-5
& Top-1 & Top-5\\
\midrule
ResNet-18
& 76.88 & 93.56 & 77.33 & 93.85 & 76.19 & 93.74\\
ResNet-20
& 68.19 & 91.04 & 68.07 & 90.97 & 68.34 & 91.25\\
ResNet-50
& 75.96 & 93.73 & 79.11 & 94.84 & 77.82 & 94.40\\
ResNet-56
& 70.54 & 91.91 & 71.44 & 92.01 & 71.81 & 92.03\\
MobileNetV2
& 70.04 & 92.54 & 70.07 & 92.55 & 71.26 & 93.01\\
\bottomrule
\end{tabular*}
\end{table}

\begin{table}[]
\centering
\caption{Classification accuracy (\%) of single-attention module models (S1–S4) using SimAM and CBAM. B0 denotes the attention-free baseline.}
\label{tab:stage-recognition}
\footnotesize
\setlength{\tabcolsep}{0pt}
\renewcommand{\arraystretch}{1.1}

\begin{tabular*}{\columnwidth}
{@{\extracolsep{\fill}}lrrrr@{}}
\toprule
& \multicolumn{2}{c}{SimAM}
& \multicolumn{2}{c}{CBAM}\\
\cmidrule(lr){2-3}
\cmidrule(lr){4-5}
Configuration
& Top-1 & Top-5
& Top-1 & Top-5\\
\midrule
B0 & 76.88 & 93.56 & 76.88 & 93.56\\
S1 & 77.99 & 93.92 & 76.73 & 93.65\\
S2 & 76.66 & 93.54 & 76.52 & 93.80\\
S3 & 77.01 & 93.47 & 76.52 & 93.45\\
S4 & 76.39 & 93.56 & 76.45 & 93.47\\
\bottomrule
\end{tabular*}
\end{table}

\begin{table}[]
\centering
\caption{Margin prediction, DAS association, and activation-state results for SimAM and CBAM. The upper five rows for each method report ALL-last results and the lower four report single-attention module methods results. $r$ and $\rho$ denote Pearson and Spearman correlations between predicted and actual margin changes. $r_\Psi$ and $\rho_\Psi$ denote correlations between DAS and actual margin change. SA is sign agreement (\%), and AC is the mean percentage of downstream activation-state changes per image.}

\label{tab:decision-analysis}

\centering
\scriptsize
\setlength{\tabcolsep}{2pt}
\renewcommand{\arraystretch}{1.1}

\resizebox{\columnwidth}{!}{%
\begin{tabular}{@{}lrrrrrrrr@{}}
\toprule
& \multicolumn{5}{c}{\textbf{Margin prediction}}
& \multicolumn{3}{c}{\textbf{DAS / activation}}\\
\cmidrule(lr){2-6}
\cmidrule(lr){7-9}
Backbone/Module
& $r$ & $\rho$ & MAE & RMSE & SA
& $r_\Psi$ & $\rho_\Psi$ & AC\\
\midrule

\multicolumn{9}{c}{\textbf{SimAM}}\\
\midrule
ResNet-18
& 1.00 & 1.00 & 0.02 & 0.03 & 99.90
& 0.97 & 0.98 & 9.39\\
ResNet-20
& 0.99 & 0.99 & 0.15 & 0.19 & 96.31
& 0.98 & 0.99 & 17.64\\
ResNet-50
& 0.92 & 0.90 & 1.91 & 2.20 & 93.62
& 0.93 & 0.91 & 18.76\\
ResNet-56
& 1.00 & 1.00 & 0.03 & 0.04 & 99.38
& 0.98 & 0.99 & 7.73\\
MobileNetV2
& 0.96 & 0.95 & 0.34 & 0.44 & 90.76
& 0.95 & 0.95 & 5.78\\
\midrule
S1
& -0.16 & -0.16 & 21.67 & 23.58 & 35.41
& -0.14 & -0.13 & 35.83\\
S2
& 0.28 & 0.28 & 7.45 & 9.37 & 85.76
& 0.28 & 0.26 & 28.71\\
S3
& 0.92 & 0.92 & 0.45 & 0.58 & 90.63
& 0.89 & 0.90 & 5.92\\
S4
& 1.00 & 1.00 & 0.02 & 0.02 & 99.90
& 0.97 & 0.97 & 7.91\\
\midrule

\multicolumn{9}{c}{\textbf{CBAM}}\\
\midrule
ResNet-18
& 1.00 & 1.00 & 0.02 & 0.03 & 99.99
& 0.94 & 0.97 & 6.46\\
ResNet-20
& 1.00 & 1.00 & 0.09 & 0.12 & 99.86
& 0.95 & 0.99 & 10.89\\
ResNet-50
& 1.00 & 1.00 & 0.22 & 0.29 & 99.88
& 0.92 & 0.92 & 11.37\\
ResNet-56
& 1.00 & 1.00 & 0.14 & 0.19 & 99.81
& 0.96 & 0.98 & 11.52\\
MobileNetV2
& 0.90 & 0.89 & 3.70 & 5.15 & 92.62
& 0.87 & 0.87 & 20.80\\
\midrule
S1
& -0.04 & -0.03 & 11.16 & 12.82 & 74.01
& -0.03 & -0.04 & 29.73\\
S2
& 0.68 & 0.62 & 14.39 & 15.92 & 98.74
& 0.17 & 0.17 & 31.14\\
S3
& 0.77 & 0.70 & 5.32 & 6.78 & 85.33
& 0.72 & 0.73 & 20.52\\
S4
& 1.00 & 1.00 & 0.03 & 0.06 & 99.98
& 0.93 & 0.97 & 7.95\\
\bottomrule
\end{tabular}%
}
\end{table}

\subsection{Architectures and Attention Placement}

ResNet architectures employ residual learning through skip connections~\cite{he2016deep}. ResNet-18 and ResNet-50 comprise four residual stages with $[2,2,2,2]$ BasicBlocks and $[3,4,6,3]$ Bottleneck blocks, respectively. Both are adapted to CIFAR-100 using a $3\times3$ stride-one stem without max pooling. ResNet-20 and ResNet-56 retain the three-stage CIFAR architecture, with $[3,3,3]$ and $[9,9,9]$ BasicBlocks, respectively, and stage widths of 16, 32, and 64 channels. All ResNet stems include BatchNorm and ReLU. MobileNetV2 employs inverted residual blocks and linear bottlenecks~\cite{sandler2018mobilenetv2}. The CIFAR-adapted MobileNetV2 uses stride one in the stem and the first 24-channel block, while retaining its inverted-residual structure, ReLU6 activations, and shortcut connections. All backbones use a 100-class classifier.

For ResNet architectures, attention is inserted after Conv2 and before BN2 in each block. In ResNet-50, the subsequent Bottleneck operations, including Conv3, follow the attention module. In MobileNetV2, attention is placed after the depthwise convolution and before BatchNorm and ReLU6. The same insertion boundaries are used for SimAM and CBAM, with CBAM adapted to match the SimAM placement. SimAM uses
$\lambda=10^{-4}$ and introduces no trainable parameters,
whereas CBAM parameters are optimized jointly with the backbone. The ALL configurations contain 8, 9, 16, 27, and 17 attention modules in ResNet-18, ResNet-20, ResNet-50, ResNet-56, and MobileNetV2, respectively. Figure~ \ref{fig:cifar-architecture} illustrates the backbone architectures and corresponding attention insertion positions.

\subsection{Training and Evaluation Protocol}

All models are trained from scratch using the same data split and random seed policy. For each backbone, identical initializations are used for the corresponding backbone parameters across attention configurations. We employ stochastic gradient descent (SGD) with a batch size of 128, momentum of 0.9, and weight decay of $5\times10^{-4}$.

Training proceeds for 64,000 optimizer updates, starting with a learning rate of 0.1, which is reduced by a factor of ten after 32,000 and 48,000 updates. Data augmentation includes four-pixel zero-padding followed by random $32\times32$ cropping and horizontal flipping. RGB channels are normalized using means $(0.4914,0.4822,0.4465)$ and standard deviations
$(0.2023,0.1994,0.2010)$. Validation and test images undergo
the same normalization without augmentation.

For each configuration, the checkpoint with the highest validation Top-1 accuracy is selected, with ties resolved in favor of the earliest epoch. The selected checkpoint is evaluated on the test set using Top-1 and Top-5 accuracy.

\subsection{Decision-Space Analysis}

Following Section~\ref{sec:methodology}, we use the ground-truth class as the target and select the strongest non-target class from the bypass logits as its competitor. This target--competitor pair is fixed for both the predicted and actual margin-change calculations, ensuring that the two measurements refer to the same decision boundary. For ALL-trained models, the reference evaluation bypasses only the final attention module, while all preceding attention modules remain active in both evaluations. Thus, the comparison isolates the effect of the final module conditioned on the preceding attention operations. In S1--S4, the analyzed module is the only attention operation in the network.

The margin gradient is evaluated at the pre-attention feature by backpropagating through the downstream network, with the selected attention module excluded and residual shortcuts held fixed. We also record changes in downstream ReLU and ReLU6 activation states between the reference and attended evaluations. The gradient projection provides a local first-order prediction of the margin change. It is exact when the feature displacement remains within the same downstream affine region; otherwise, activation-state changes may contribute to the prediction error. To examine the relationship between DAS and the actual margin change, we sort test samples by DAS and divide them into ten equal-count groups. For each group, we compute the mean DAS and the mean actual margin change, representing the group by a single point. This analysis reveals how the average margin response varies with the degree of directional alignment.

\newcommand{\finalStudyRowLabel}[1]{%
\begin{minipage}[c]{0.055\textwidth}
\centering
\rotatebox{90}{\footnotesize\bfseries #1}
\end{minipage}%
}

\newcommand{\pairedfigspace}{\vspace{5mm}}

\begin{figure*}[]
\centering

\noindent
\finalStudyRowLabel{SimAM}\hfill%
\begin{minipage}[c]{0.935\textwidth}
\centering
\begin{minipage}[c]{0.192\linewidth}
\centering
\includegraphics[width=\linewidth]{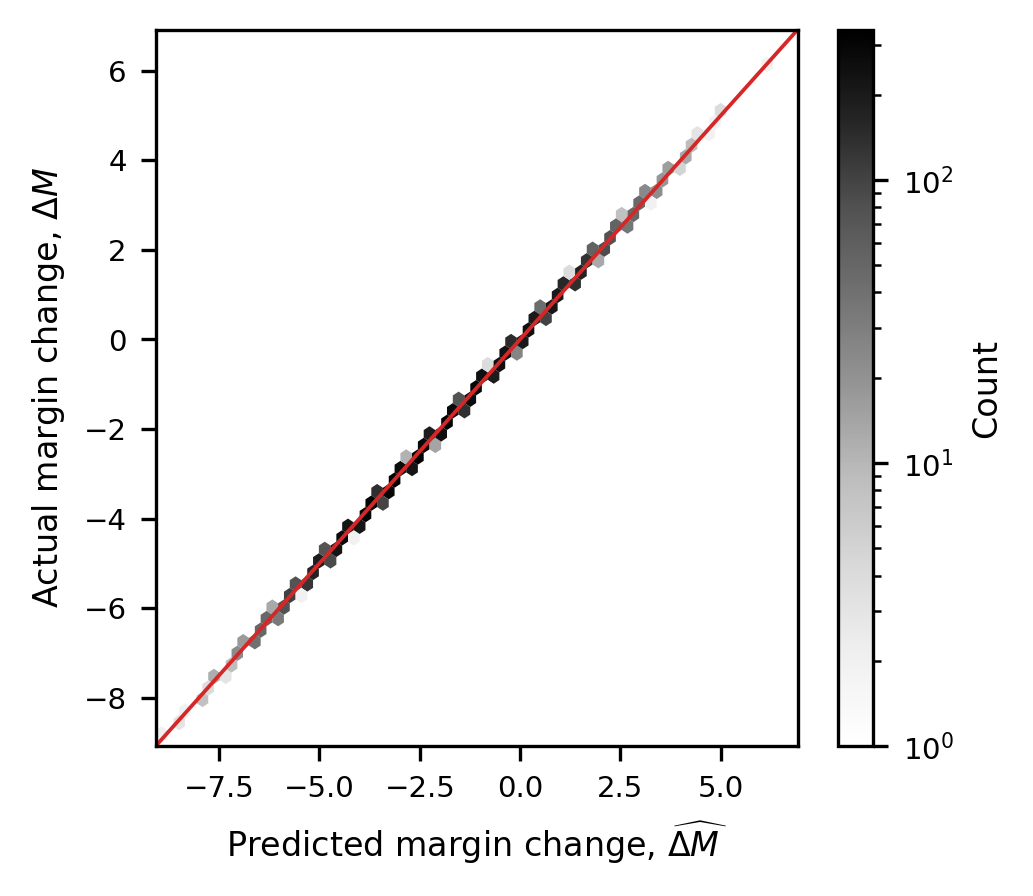}
\par\vspace{-1pt}
{\scriptsize ResNet-18}
\end{minipage}\hfill%
\begin{minipage}[c]{0.192\linewidth}
\centering
\includegraphics[width=\linewidth]{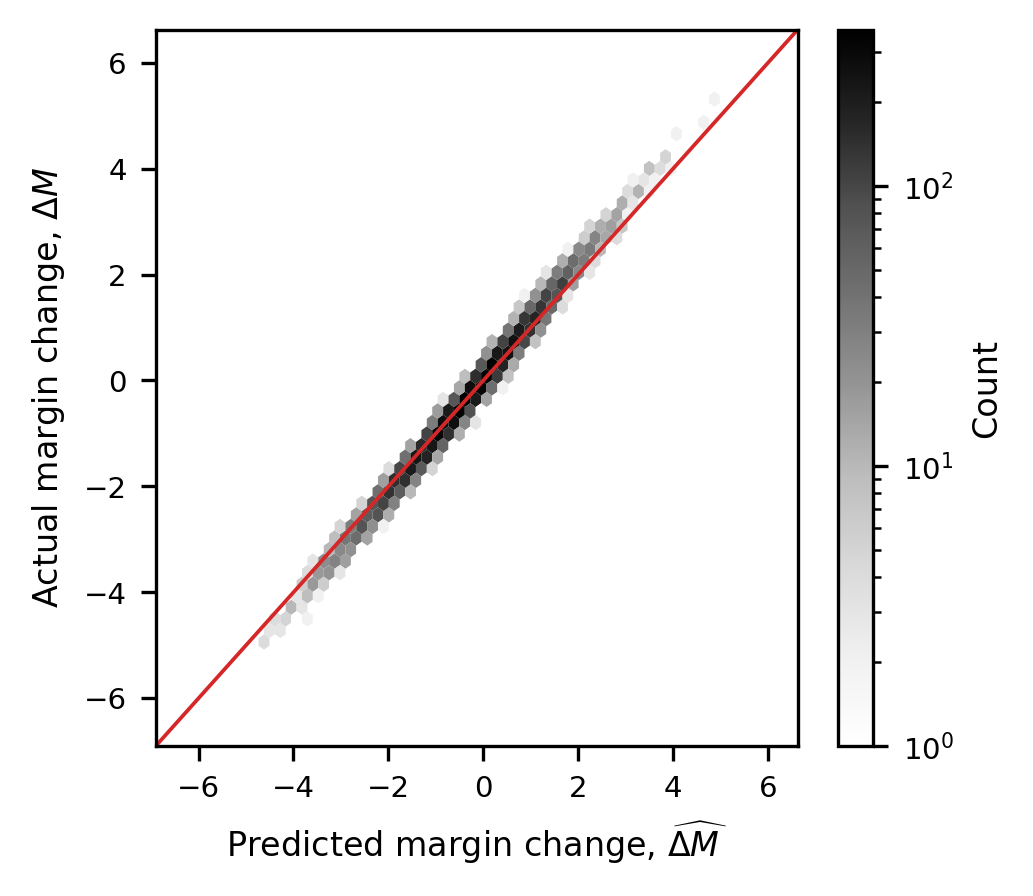}
\par\vspace{-1pt}
{\scriptsize ResNet-20}
\end{minipage}\hfill%
\begin{minipage}[c]{0.192\linewidth}
\centering
\includegraphics[width=\linewidth]{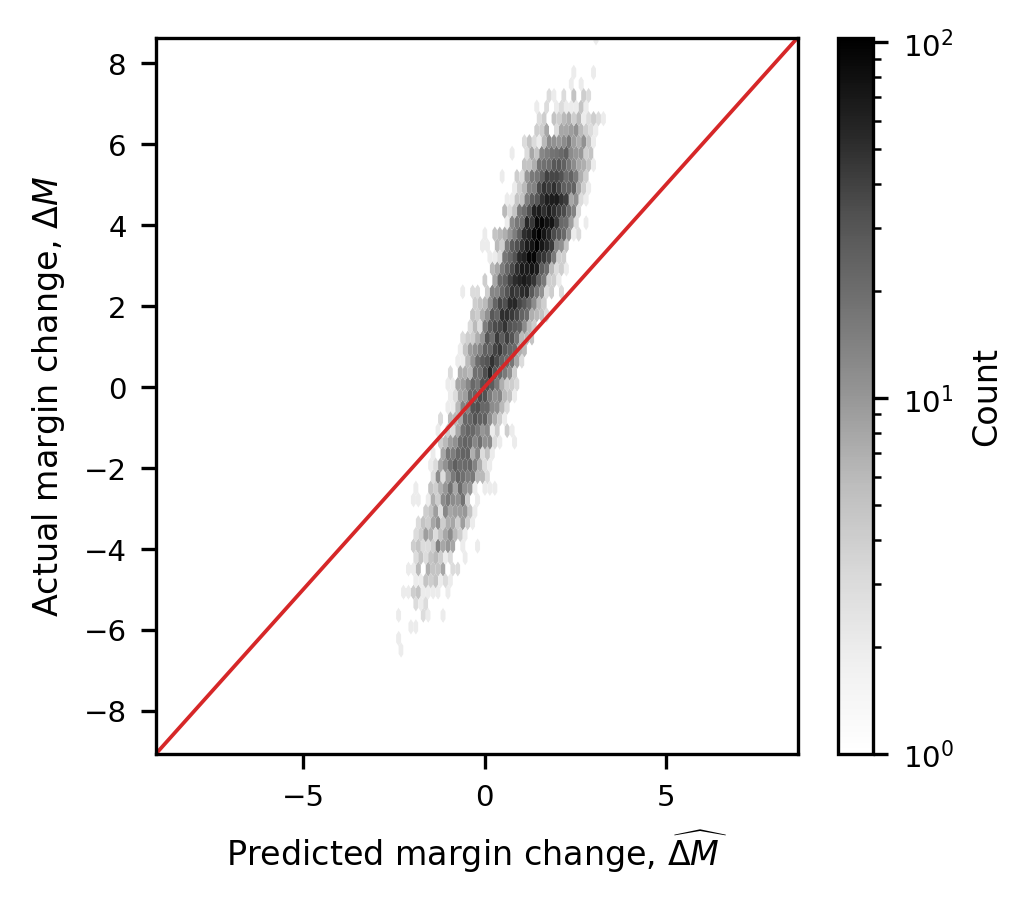}
\par\vspace{-1pt}
{\scriptsize ResNet-50}
\end{minipage}\hfill%
\begin{minipage}[c]{0.192\linewidth}
\centering
\includegraphics[width=\linewidth]{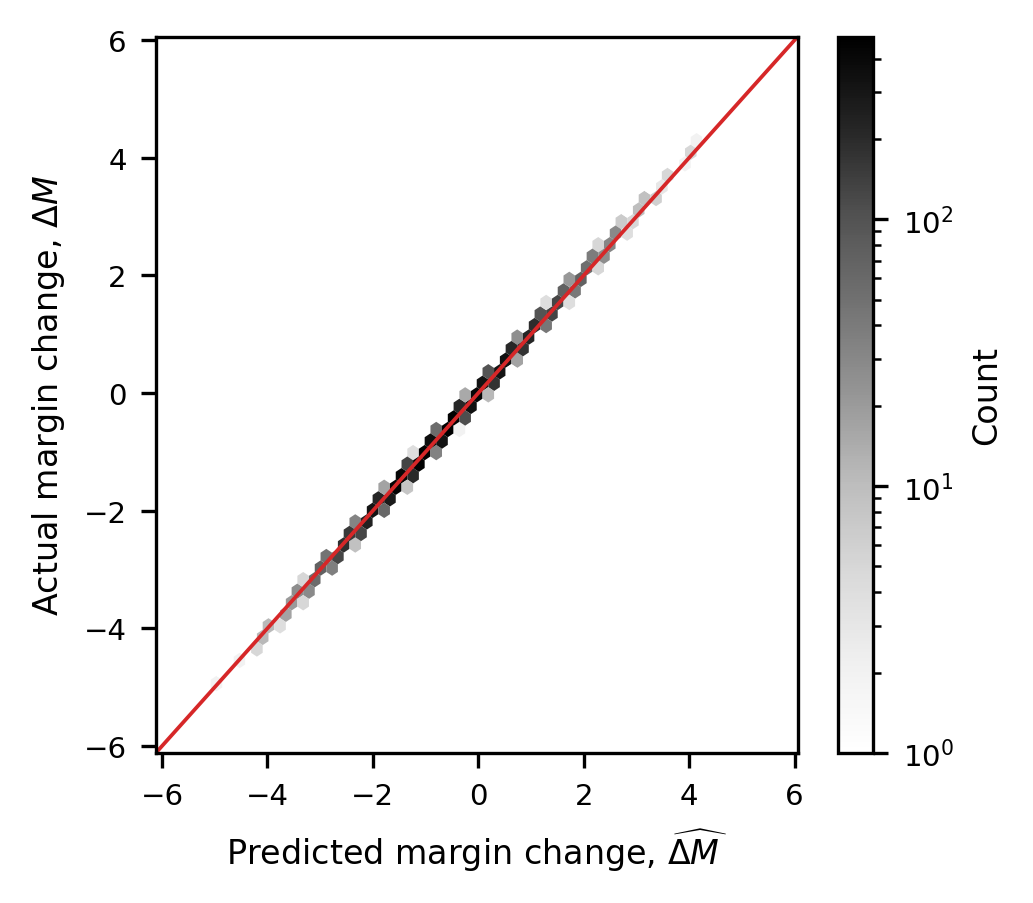}
\par\vspace{-1pt}
{\scriptsize ResNet-56}
\end{minipage}\hfill%
\begin{minipage}[c]{0.192\linewidth}
\centering
\includegraphics[width=\linewidth]{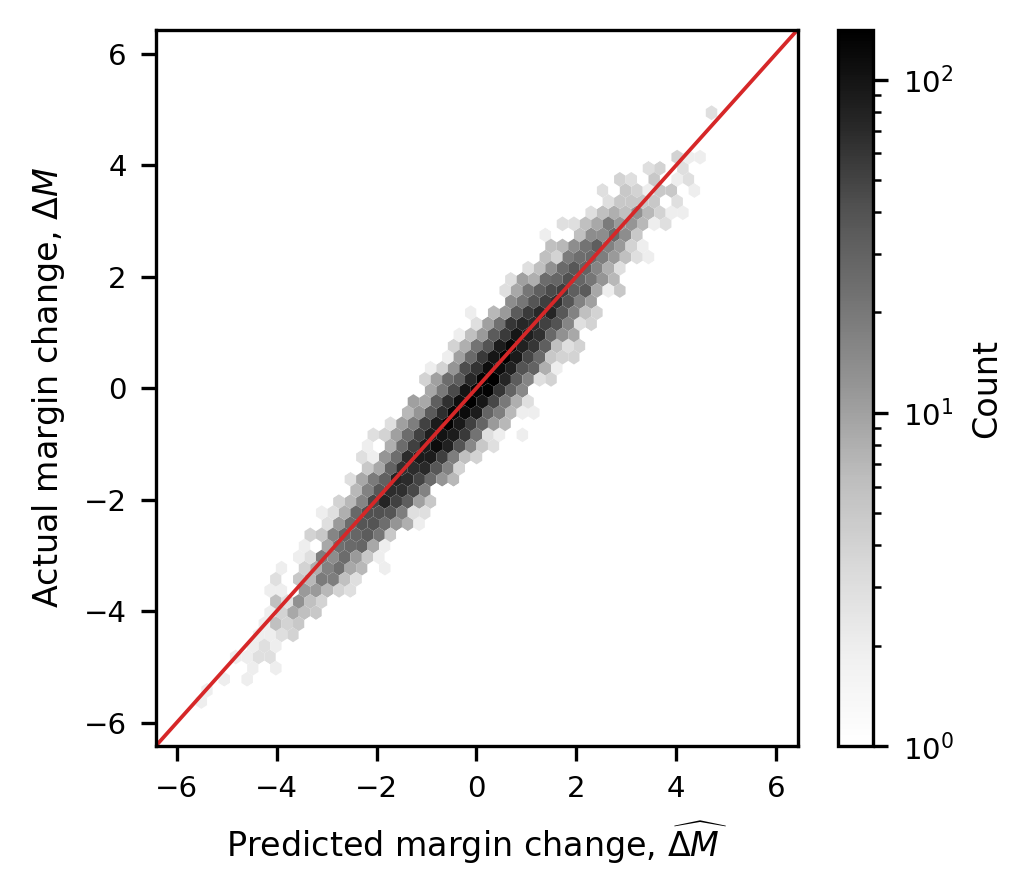}
\par\vspace{-1pt}
{\scriptsize MobileNetV2}
\end{minipage}%
\end{minipage}\par

\vspace{2mm}

\noindent
\finalStudyRowLabel{CBAM}\hfill%
\begin{minipage}[c]{0.935\textwidth}
\centering
\begin{minipage}[c]{0.192\linewidth}
\centering
\includegraphics[width=\linewidth]{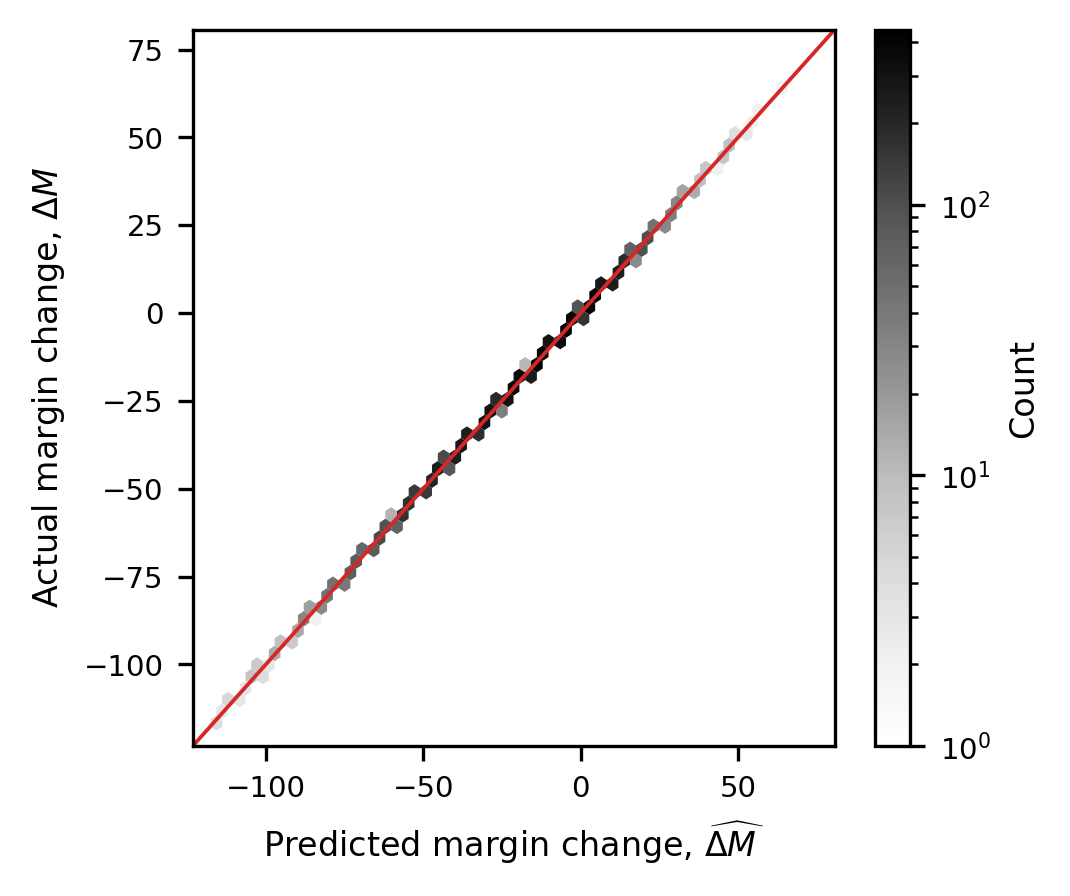}
\par\vspace{-1pt}
{\scriptsize ResNet-18}
\end{minipage}\hfill%
\begin{minipage}[c]{0.192\linewidth}
\centering
\includegraphics[width=\linewidth]{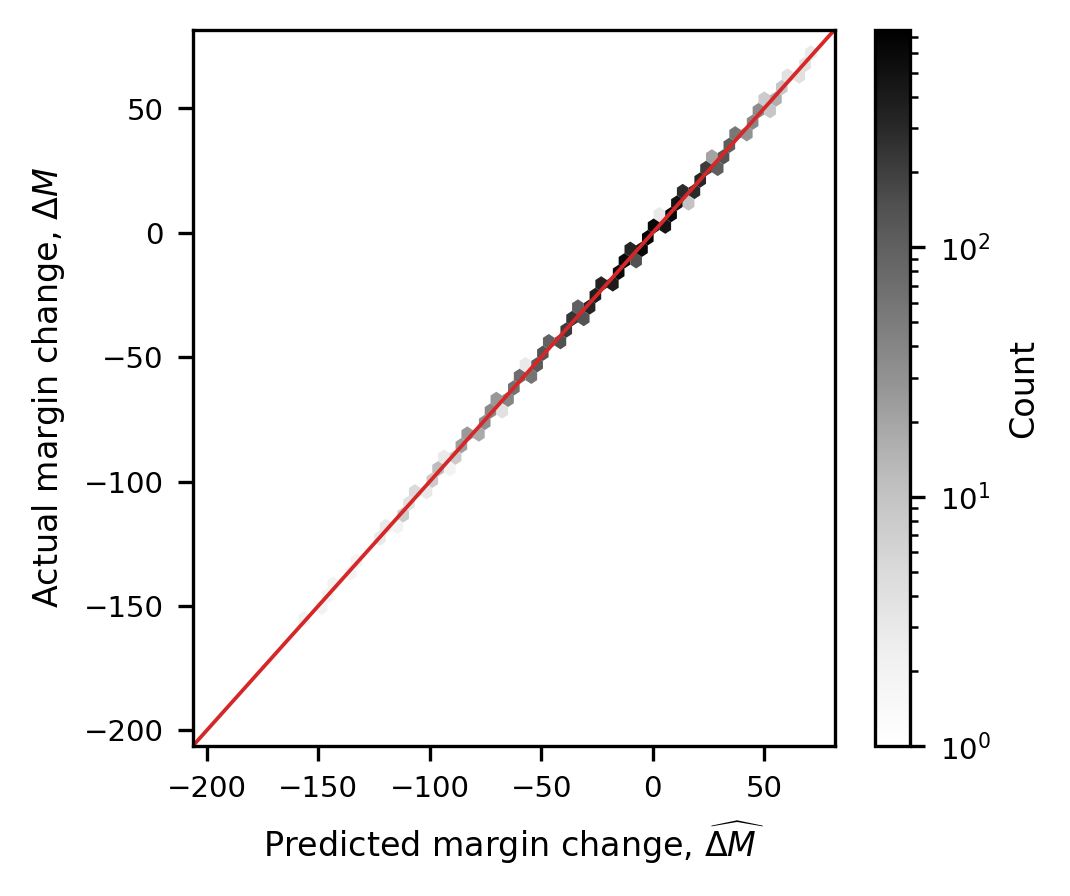}
\par\vspace{-1pt}
{\scriptsize ResNet-20}
\end{minipage}\hfill%
\begin{minipage}[c]{0.192\linewidth}
\centering
\includegraphics[width=\linewidth]{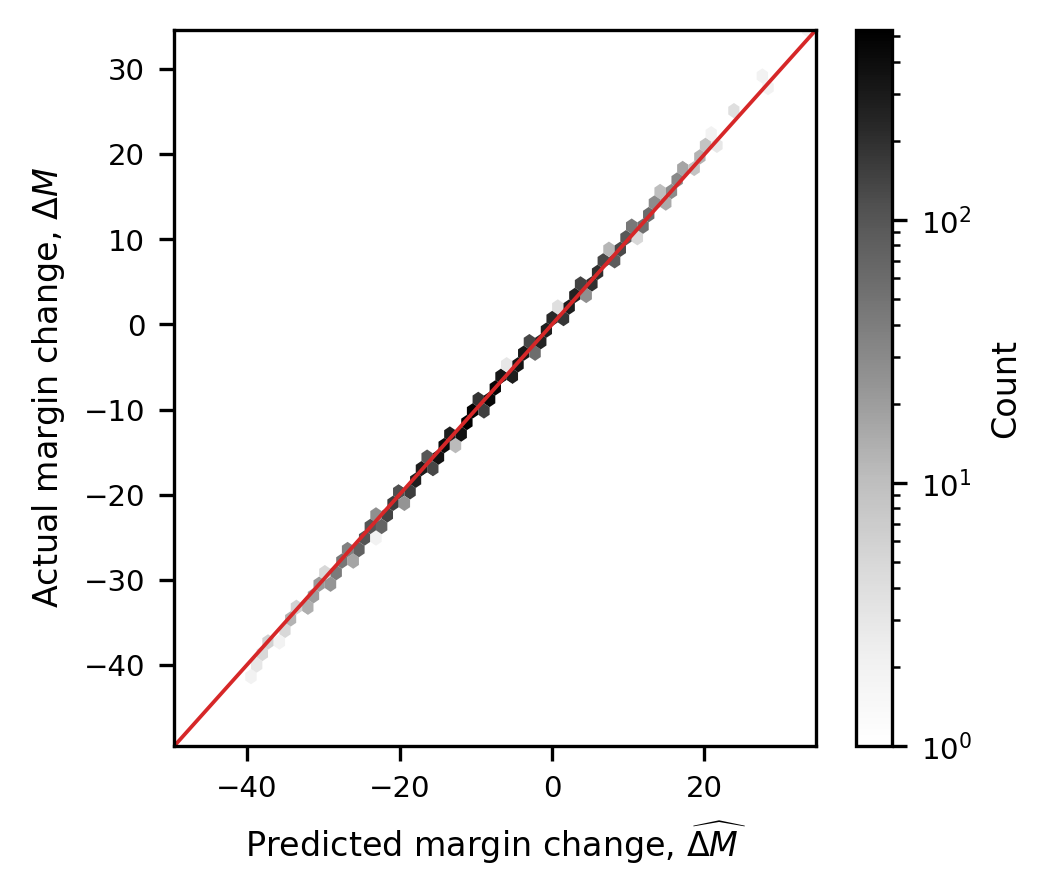}
\par\vspace{-1pt}
{\scriptsize ResNet-50}
\end{minipage}\hfill%
\begin{minipage}[c]{0.192\linewidth}
\centering
\includegraphics[width=\linewidth]{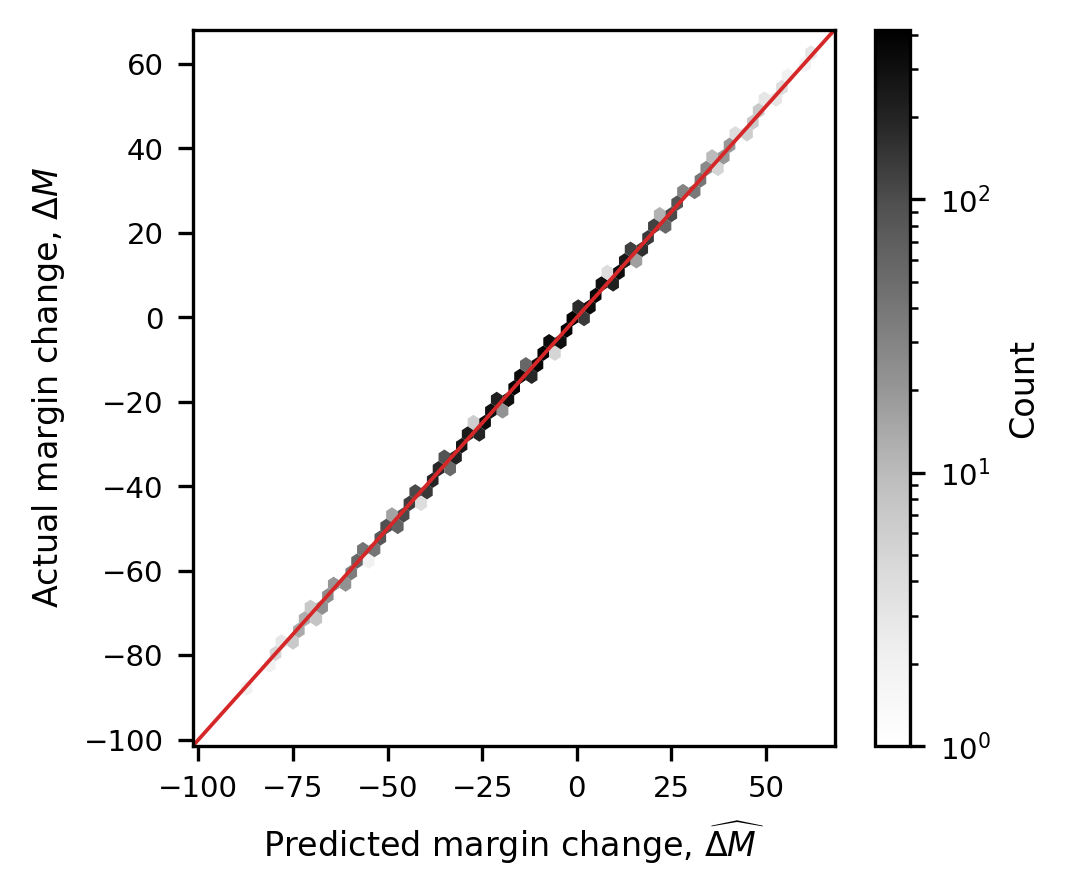}
\par\vspace{-1pt}
{\scriptsize ResNet-56}
\end{minipage}\hfill%
\begin{minipage}[c]{0.192\linewidth}
\centering
\includegraphics[width=\linewidth]{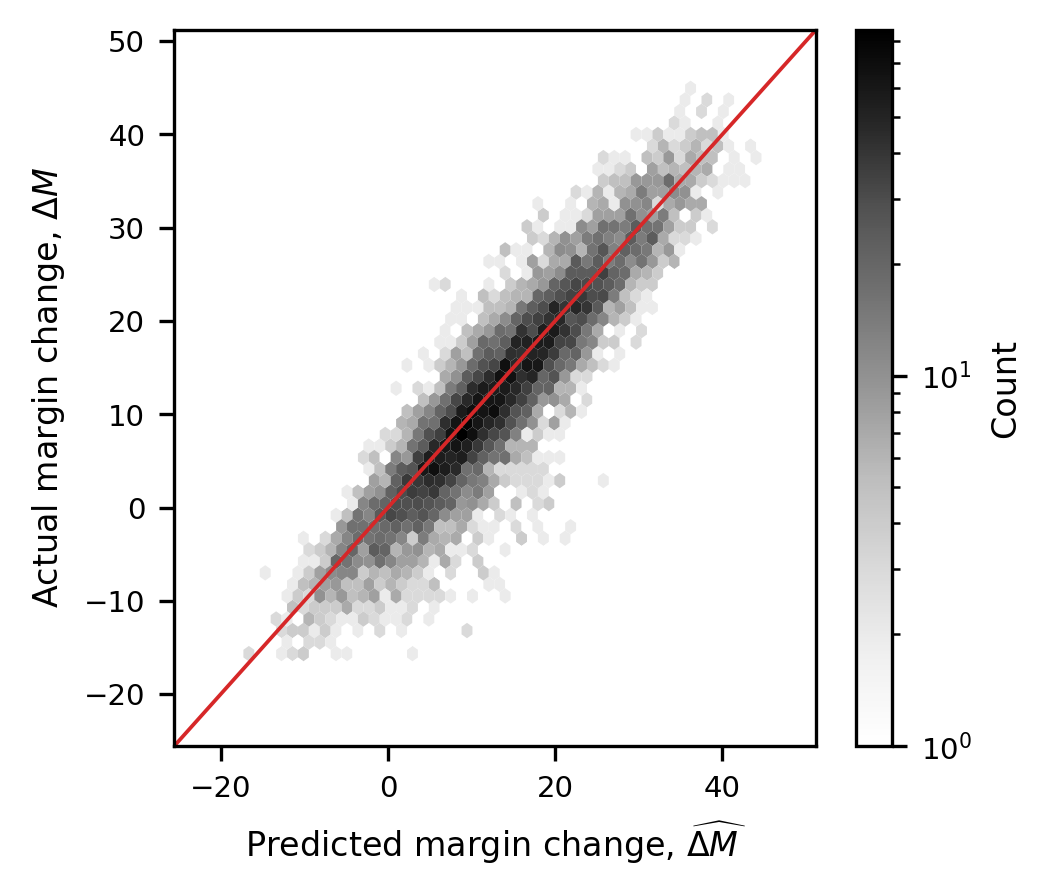}
\par\vspace{-1pt}
{\scriptsize MobileNetV2}
\end{minipage}%
\end{minipage}\par

\vspace{1mm}
\caption{Predicted versus actual margin changes for the final attention modules in ALL-trained models across five CNN backbones. Results are shown separately for SimAM and CBAM. The red line indicates exact agreement.}
\label{fig:cifar-prediction-all}

\pairedfigspace

\noindent
\finalStudyRowLabel{SimAM}\hfill%
\begin{minipage}[c]{0.935\textwidth}
\centering
\begin{minipage}[c]{0.205\linewidth}
\centering
\includegraphics[width=\linewidth]{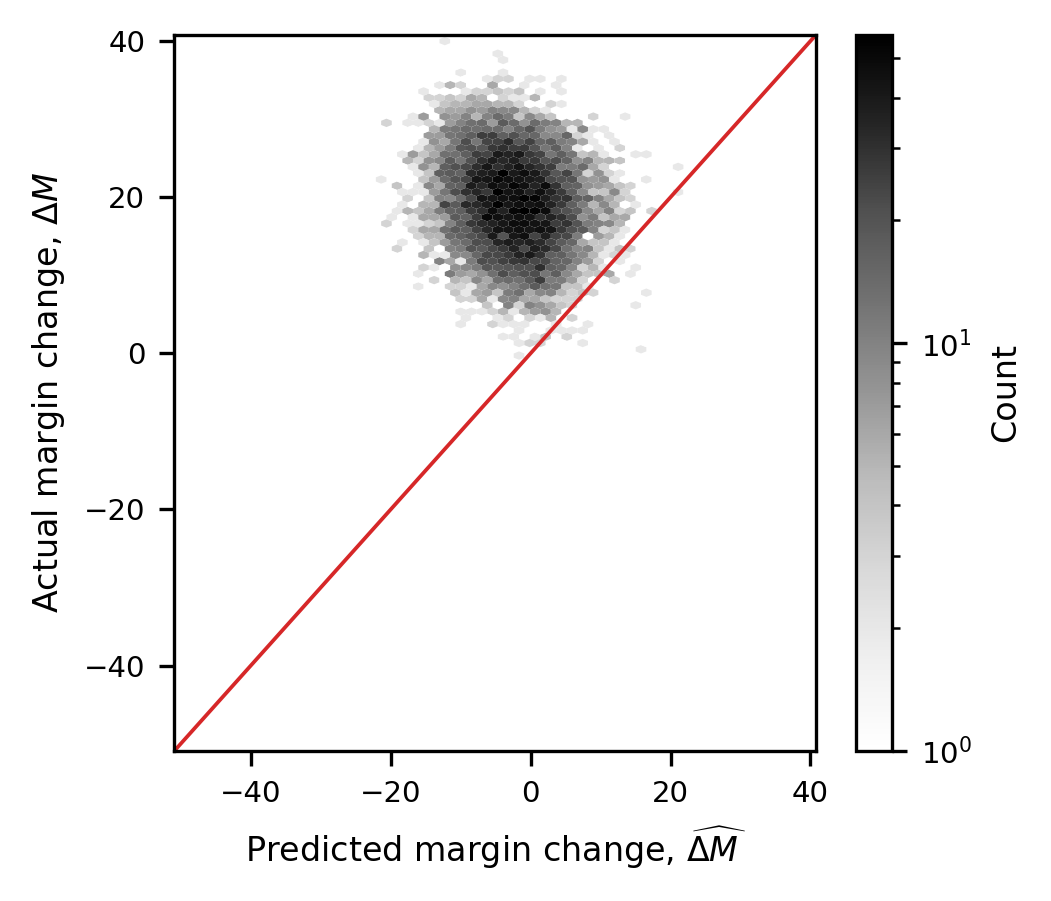}
\par\vspace{-1pt}
{\scriptsize S1}
\end{minipage}\hfill%
\begin{minipage}[c]{0.205\linewidth}
\centering
\includegraphics[width=\linewidth]{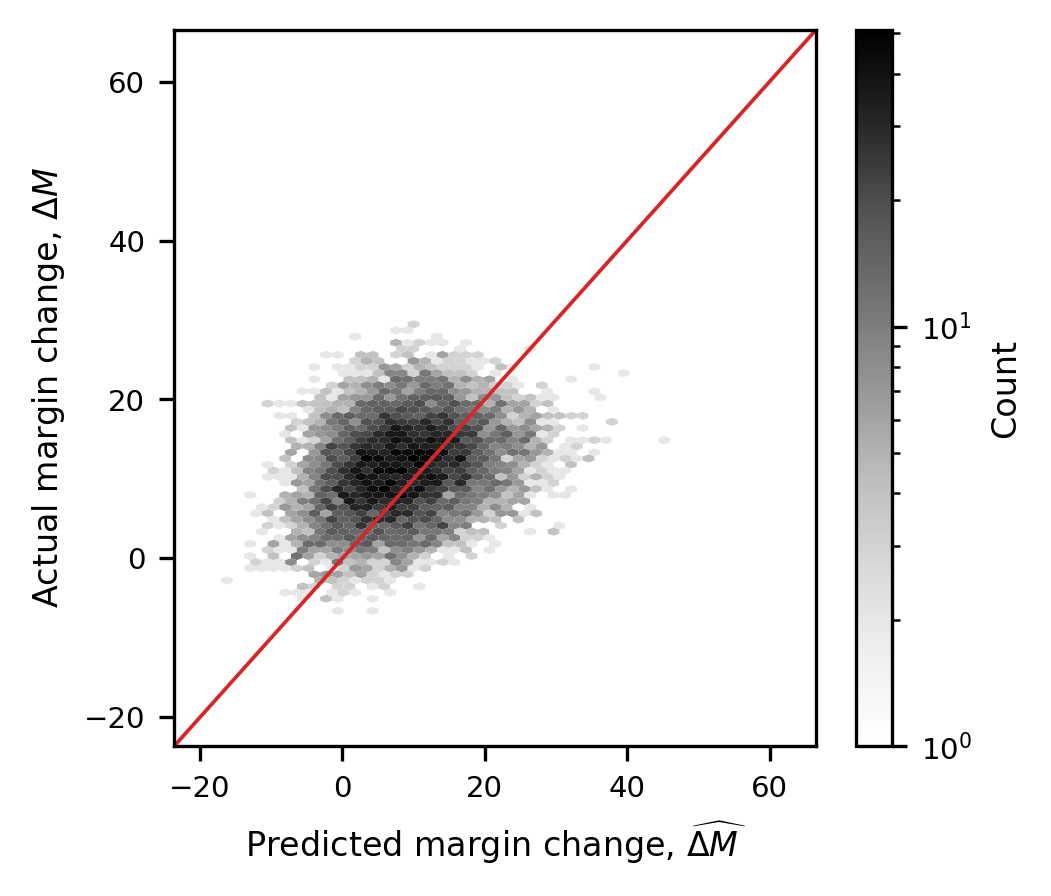}
\par\vspace{-1pt}
{\scriptsize S2}
\end{minipage}\hfill%
\begin{minipage}[c]{0.205\linewidth}
\centering
\includegraphics[width=\linewidth]{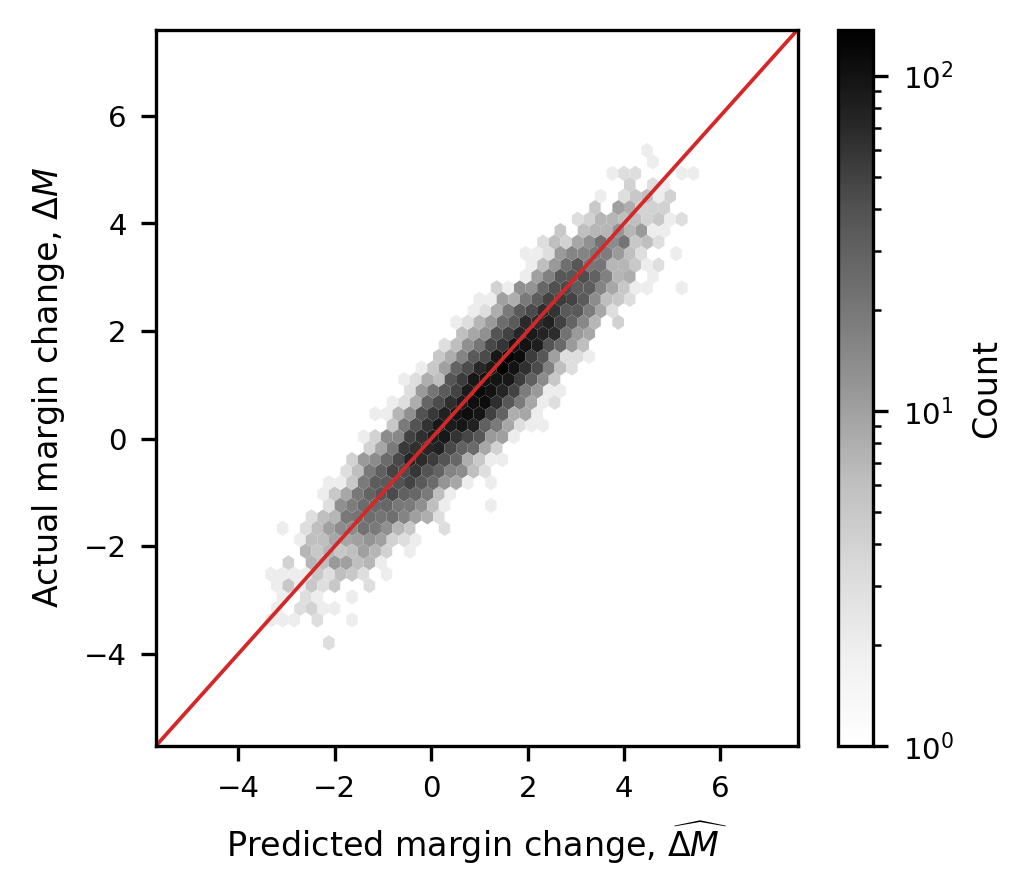}
\par\vspace{-1pt}
{\scriptsize S3}
\end{minipage}\hfill%
\begin{minipage}[c]{0.205\linewidth}
\centering
\includegraphics[width=\linewidth]{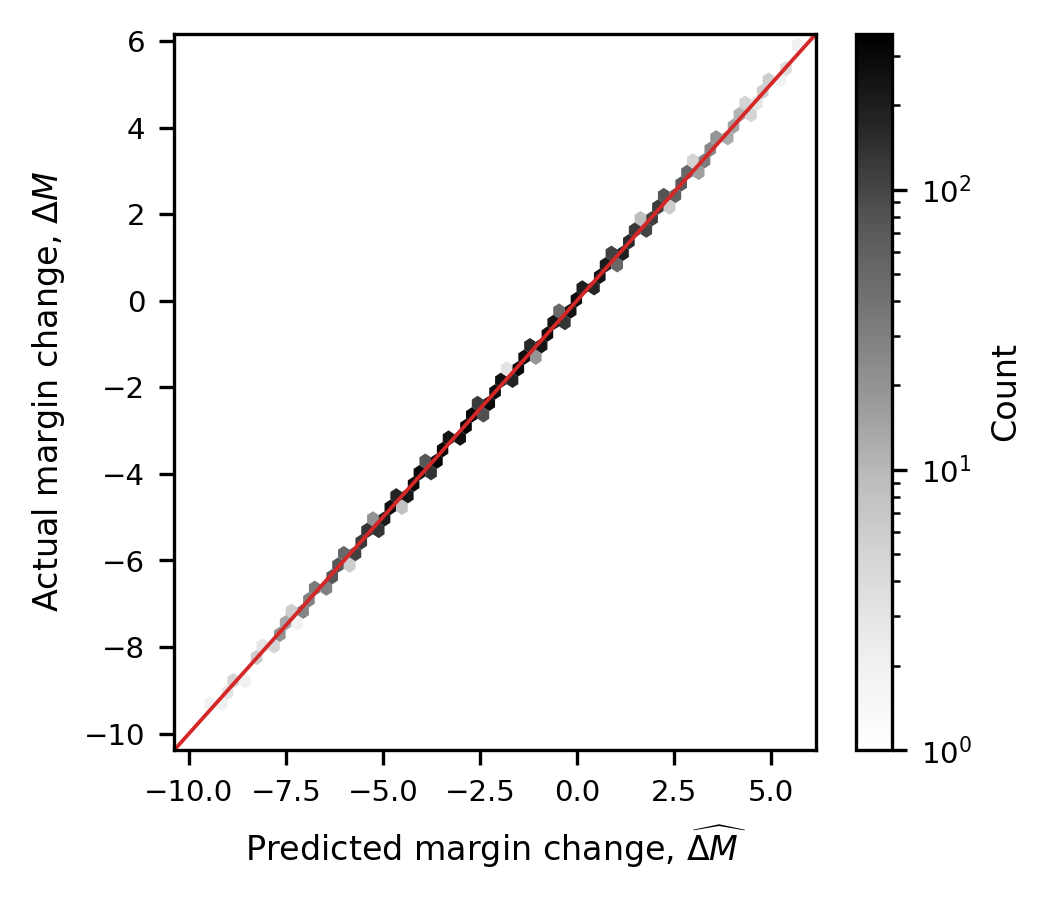}
\par\vspace{-1pt}
{\scriptsize S4}
\end{minipage}%
\end{minipage}\par

\vspace{2mm}

\noindent
\finalStudyRowLabel{CBAM}\hfill%
\begin{minipage}[c]{0.935\textwidth}
\centering
\begin{minipage}[c]{0.205\linewidth}
\centering
\includegraphics[width=\linewidth]{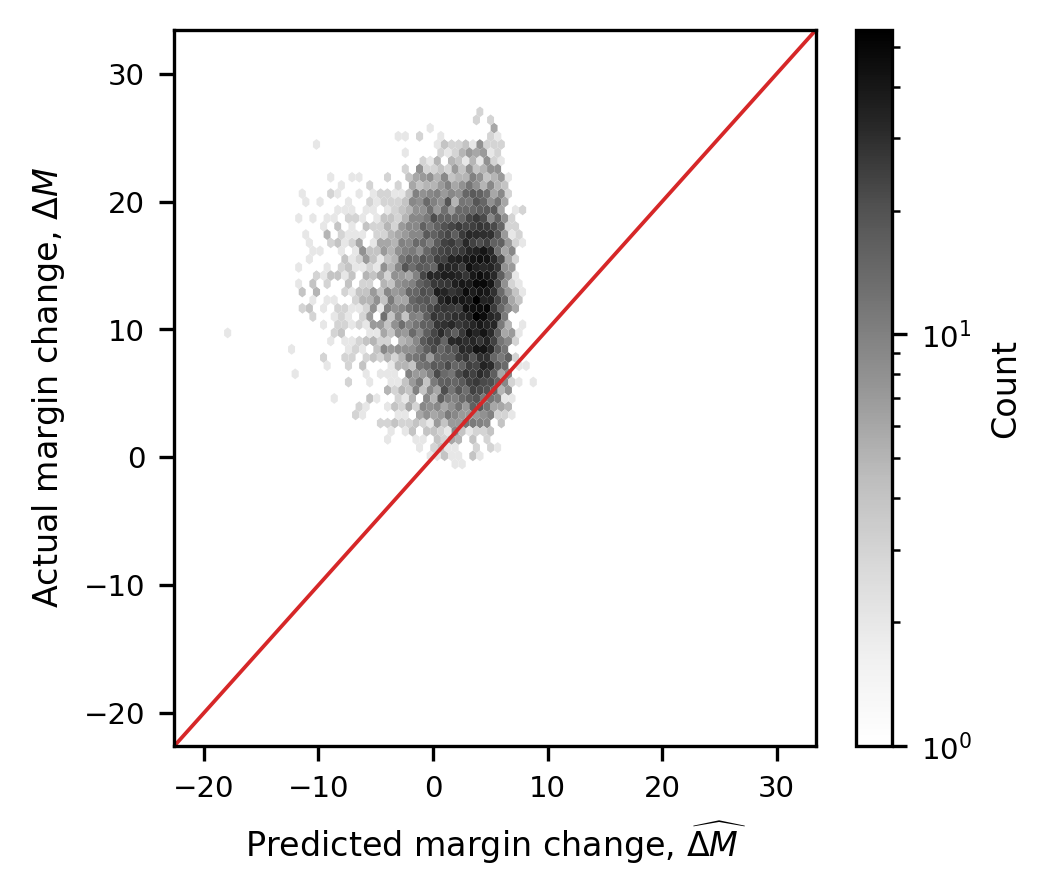}
\par\vspace{-1pt}
{\scriptsize S1}
\end{minipage}\hfill%
\begin{minipage}[c]{0.205\linewidth}
\centering
\includegraphics[width=\linewidth]{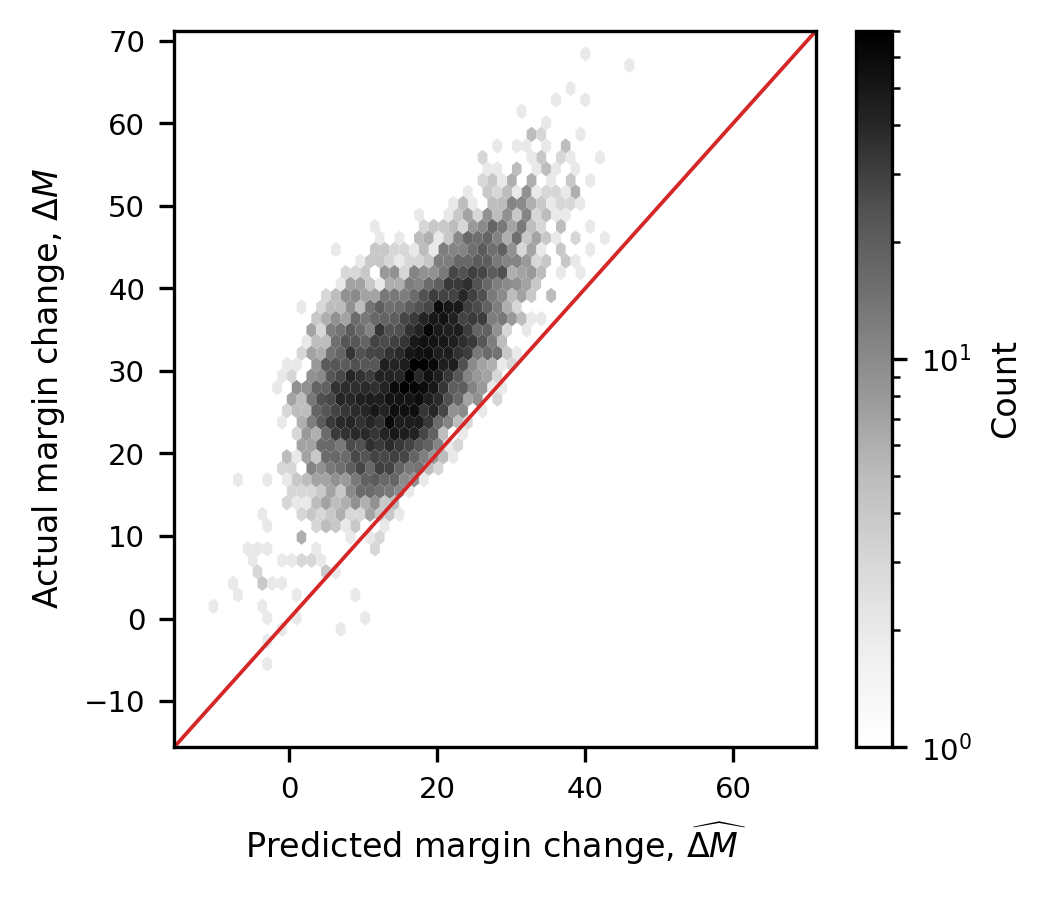}
\par\vspace{-1pt}
{\scriptsize S2}
\end{minipage}\hfill%
\begin{minipage}[c]{0.205\linewidth}
\centering
\includegraphics[width=\linewidth]{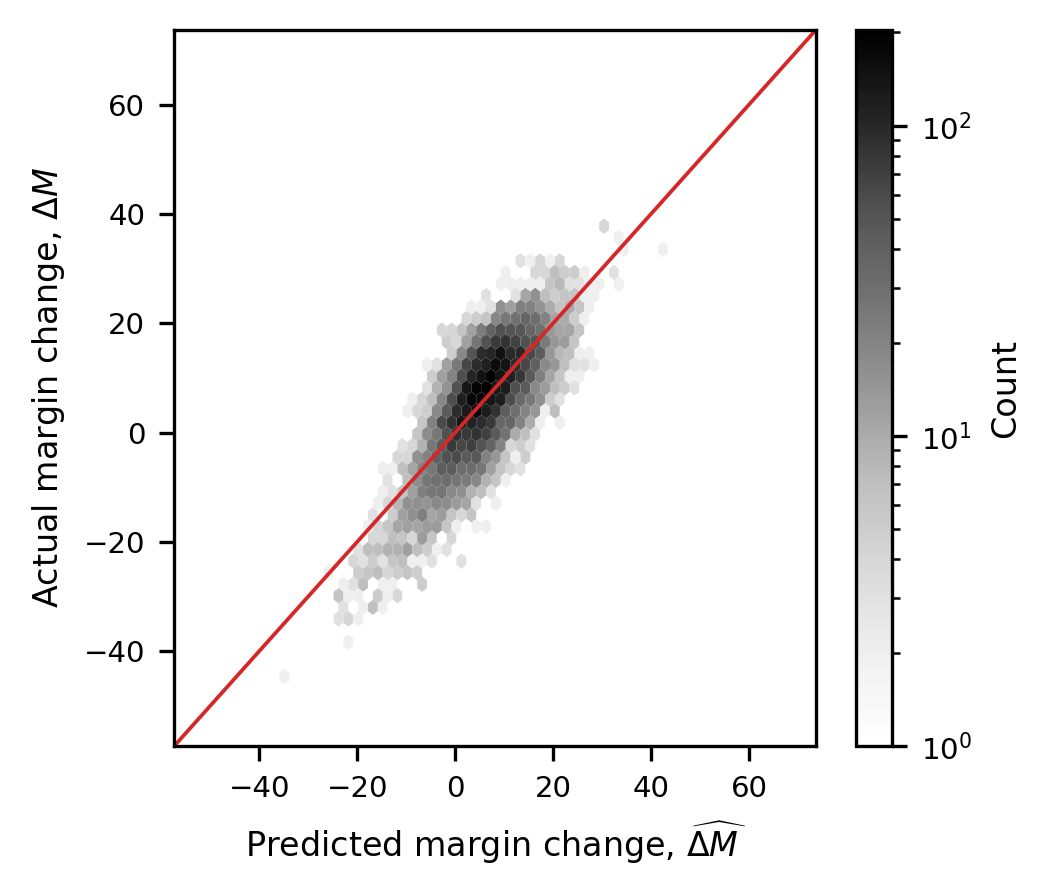}
\par\vspace{-1pt}
{\scriptsize S3}
\end{minipage}\hfill%
\begin{minipage}[c]{0.205\linewidth}
\centering
\includegraphics[width=\linewidth]{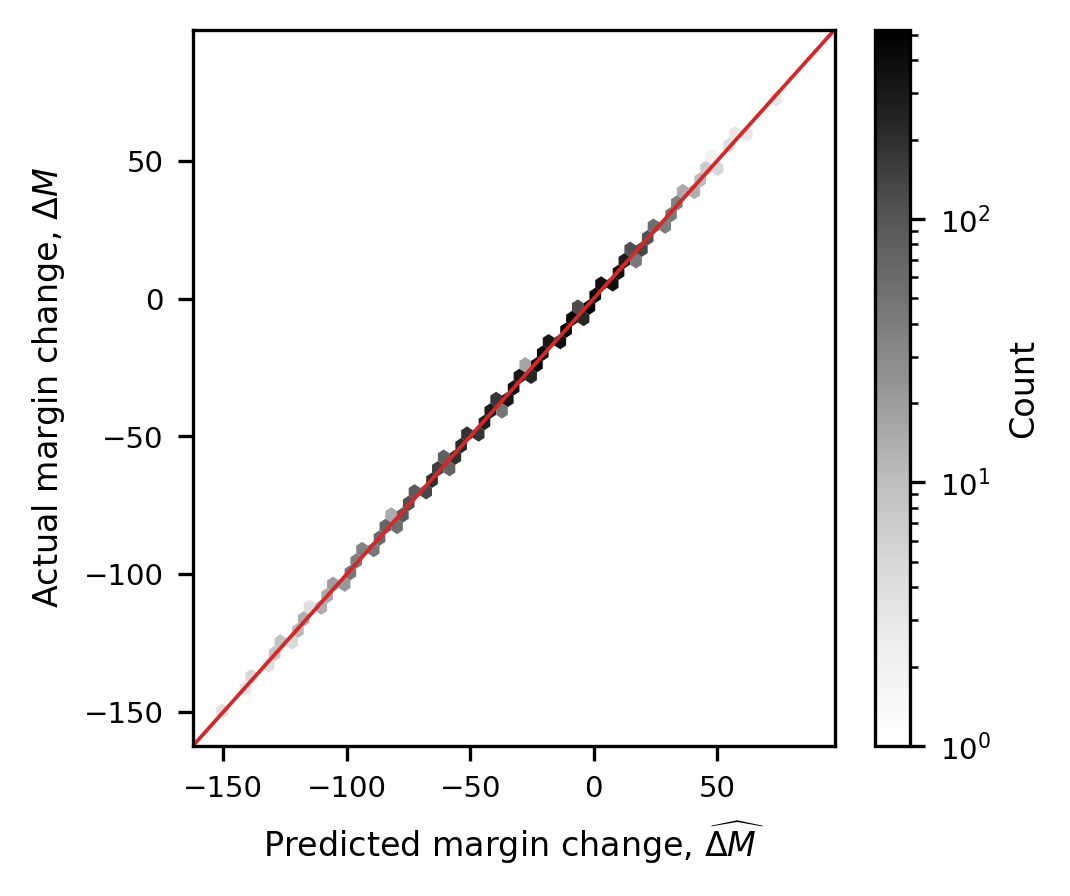}
\par\vspace{-1pt}
{\scriptsize S4}
\end{minipage}%
\end{minipage}\par

\vspace{1mm}
\caption{Predicted versus actual margin changes for independently trained single-attention module models (S1--S4) using SimAM and CBAM. The red line indicates exact agreement.}
\label{fig:cifar-prediction-single}

\end{figure*}

\begin{figure*}[]
\centering

\noindent
\finalStudyRowLabel{SimAM}\hfill%
\begin{minipage}[c]{0.935\textwidth}
\centering
\begin{minipage}[c]{0.192\linewidth}
\centering
\includegraphics[width=\linewidth]{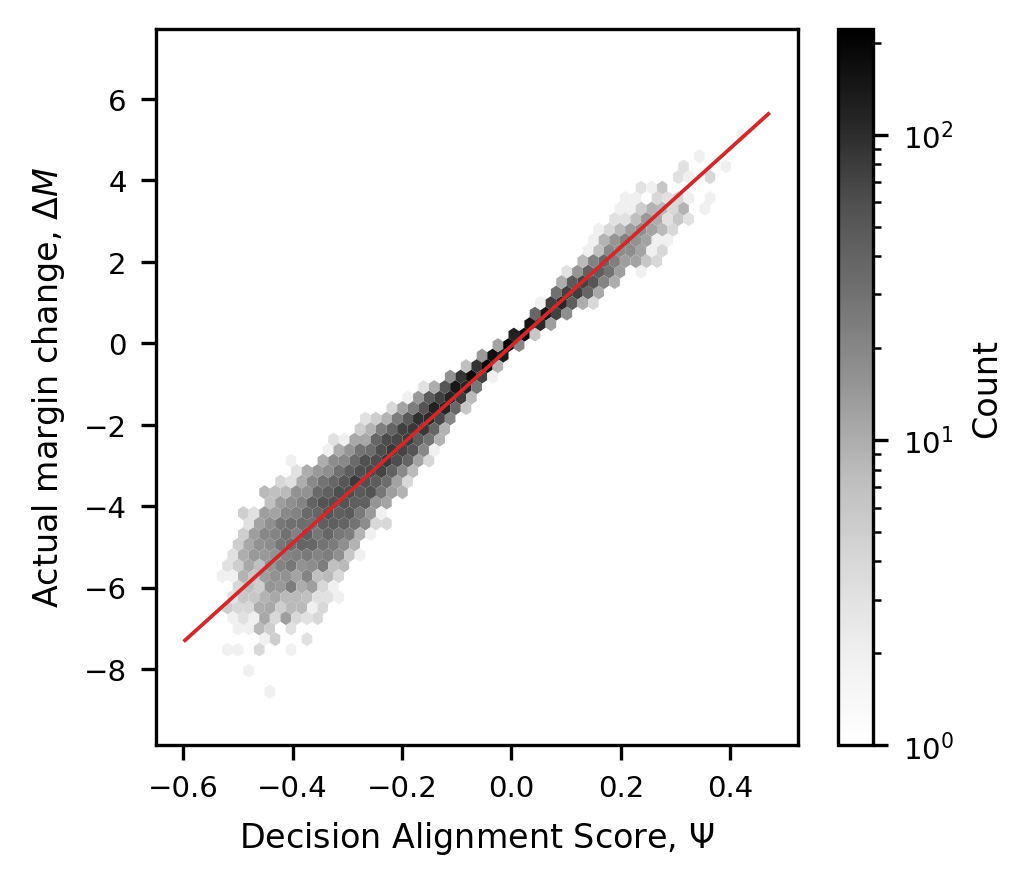}
\par\vspace{-1pt}
{\scriptsize ResNet-18}
\end{minipage}\hfill%
\begin{minipage}[c]{0.192\linewidth}
\centering
\includegraphics[width=\linewidth]{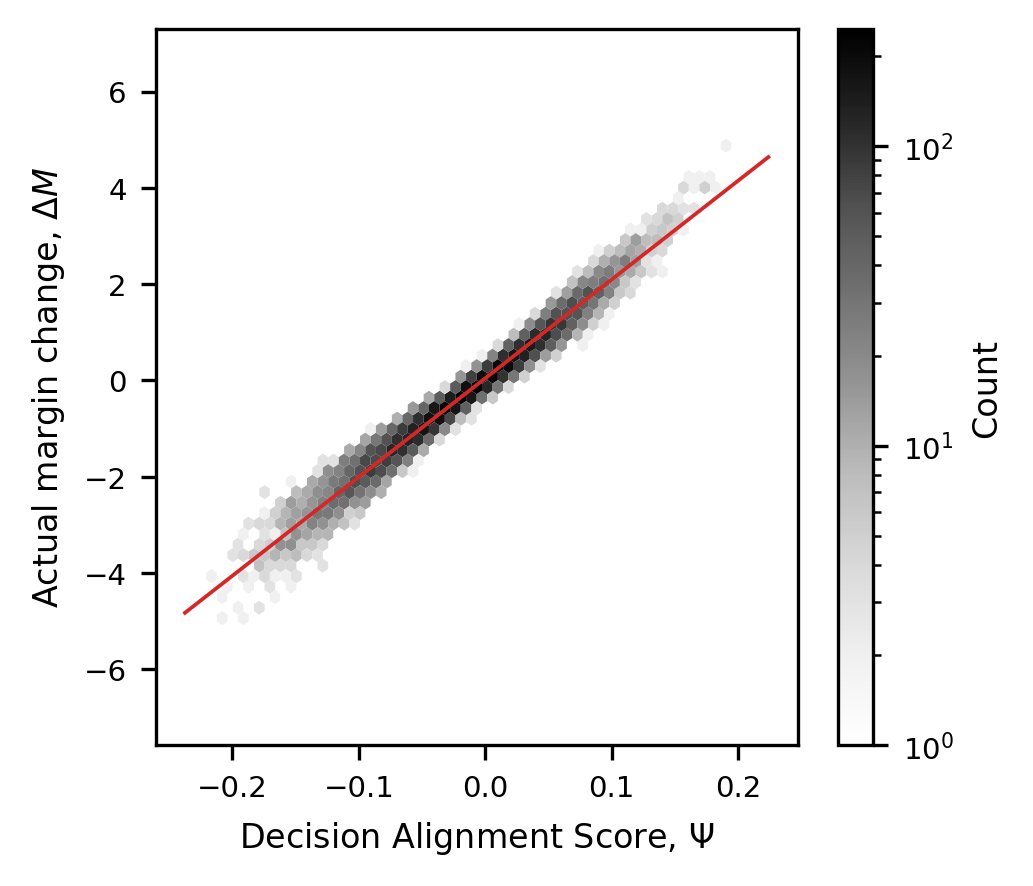}
\par\vspace{-1pt}
{\scriptsize ResNet-20}
\end{minipage}\hfill%
\begin{minipage}[c]{0.192\linewidth}
\centering
\includegraphics[width=\linewidth]{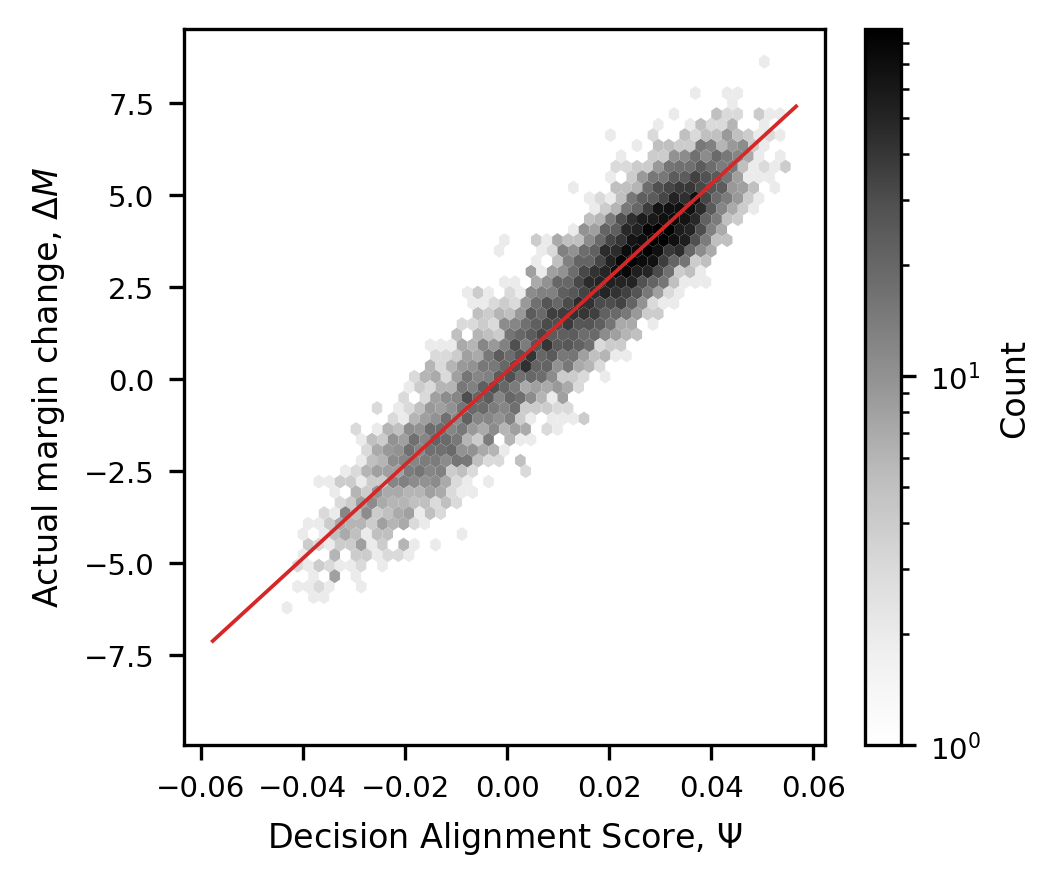}
\par\vspace{-1pt}
{\scriptsize ResNet-50}
\end{minipage}\hfill%
\begin{minipage}[c]{0.192\linewidth}
\centering
\includegraphics[width=\linewidth]{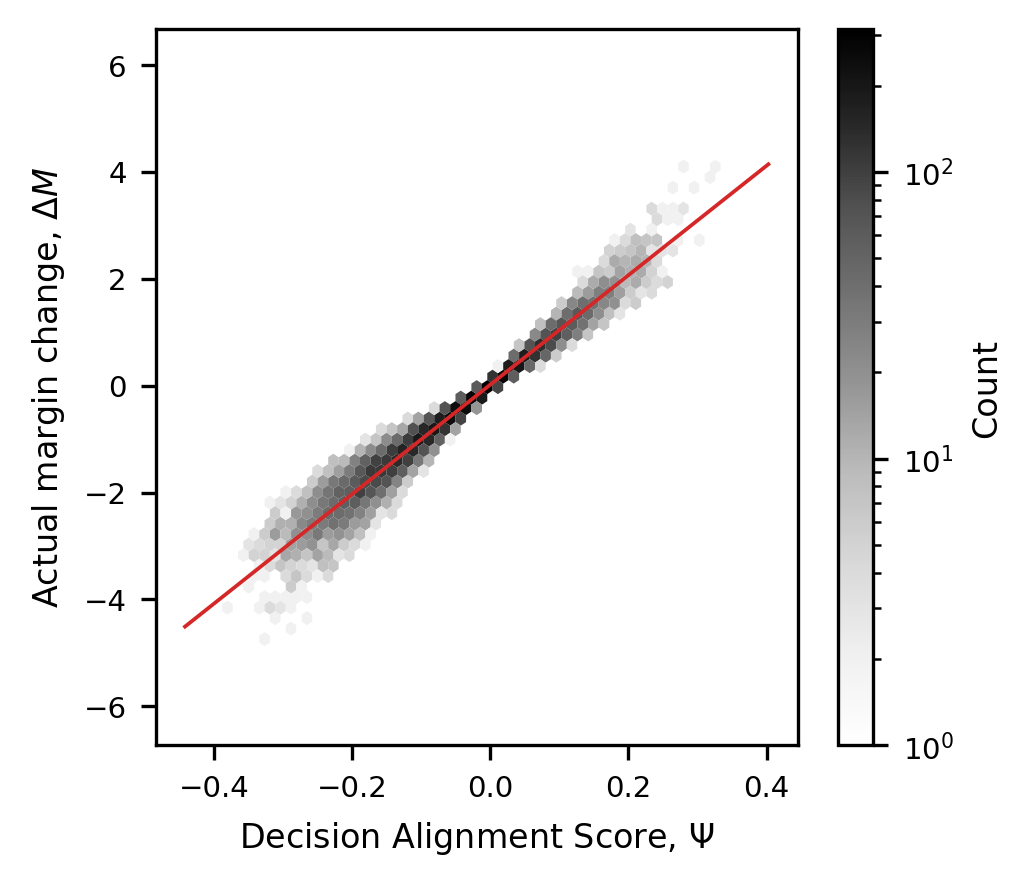}
\par\vspace{-1pt}
{\scriptsize ResNet-56}
\end{minipage}\hfill%
\begin{minipage}[c]{0.192\linewidth}
\centering
\includegraphics[width=\linewidth]{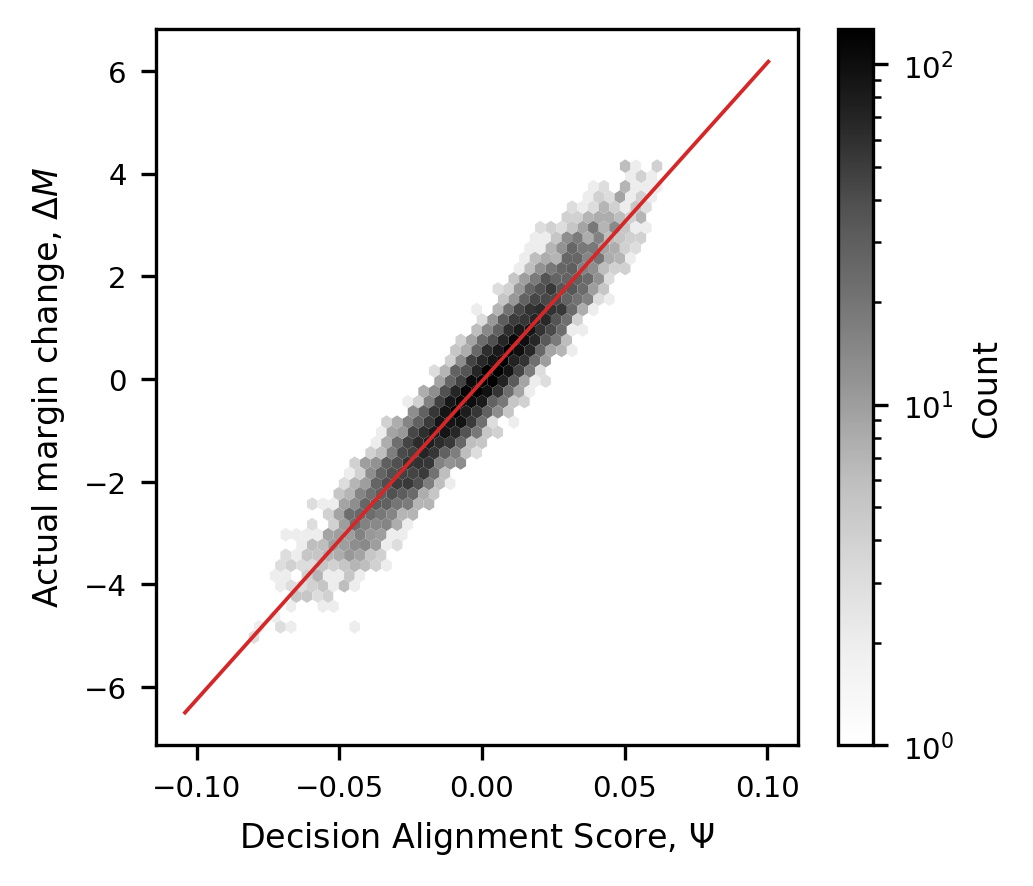}
\par\vspace{-1pt}
{\scriptsize MobileNetV2}
\end{minipage}%
\end{minipage}\par

\vspace{2mm}

\noindent
\finalStudyRowLabel{CBAM}\hfill%
\begin{minipage}[c]{0.935\textwidth}
\centering
\begin{minipage}[c]{0.192\linewidth}
\centering
\includegraphics[width=\linewidth]{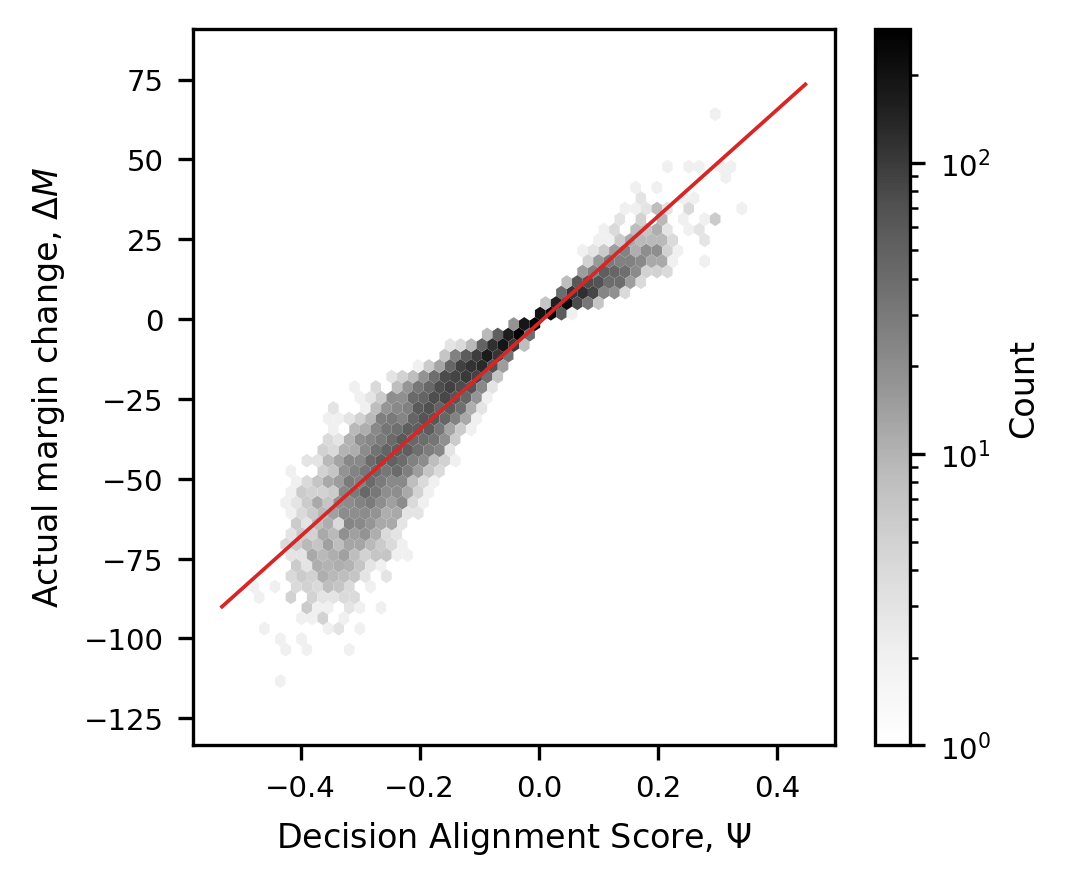}
\par\vspace{-1pt}
{\scriptsize ResNet-18}
\end{minipage}\hfill%
\begin{minipage}[c]{0.192\linewidth}
\centering
\includegraphics[width=\linewidth]{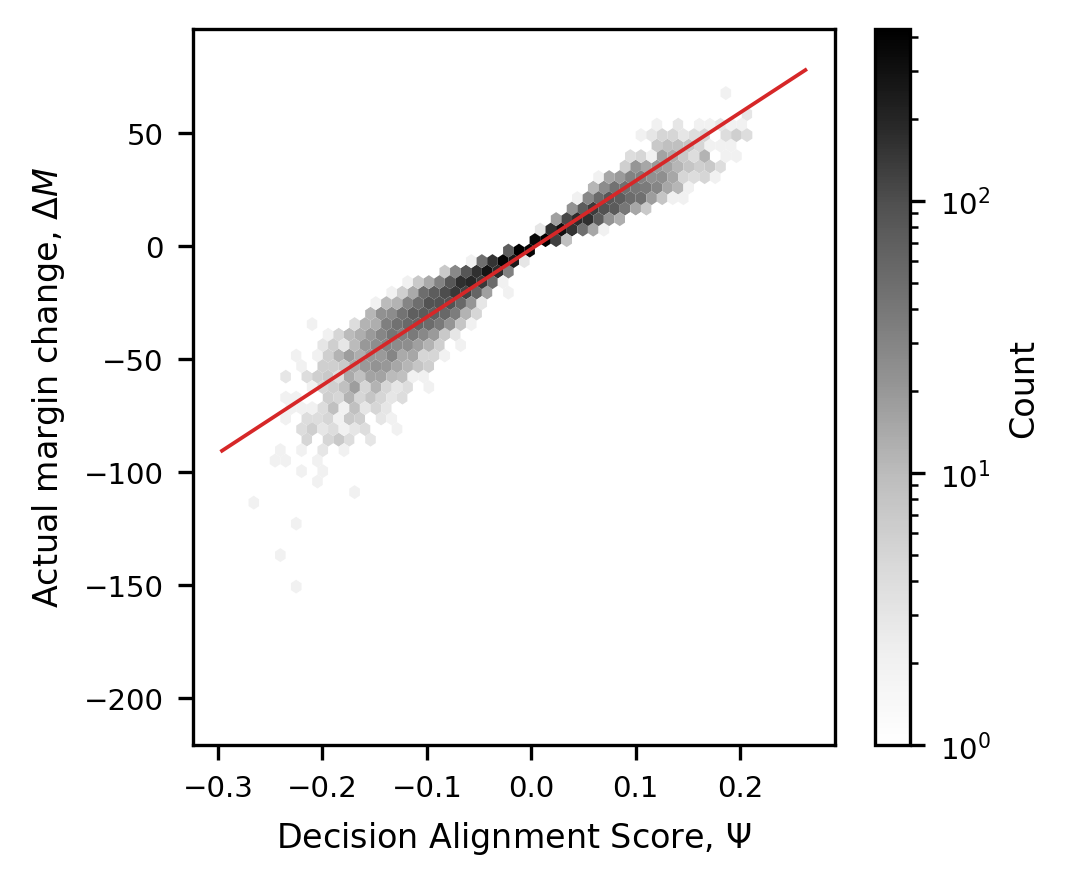}
\par\vspace{-1pt}
{\scriptsize ResNet-20}
\end{minipage}\hfill%
\begin{minipage}[c]{0.192\linewidth}
\centering
\includegraphics[width=\linewidth]{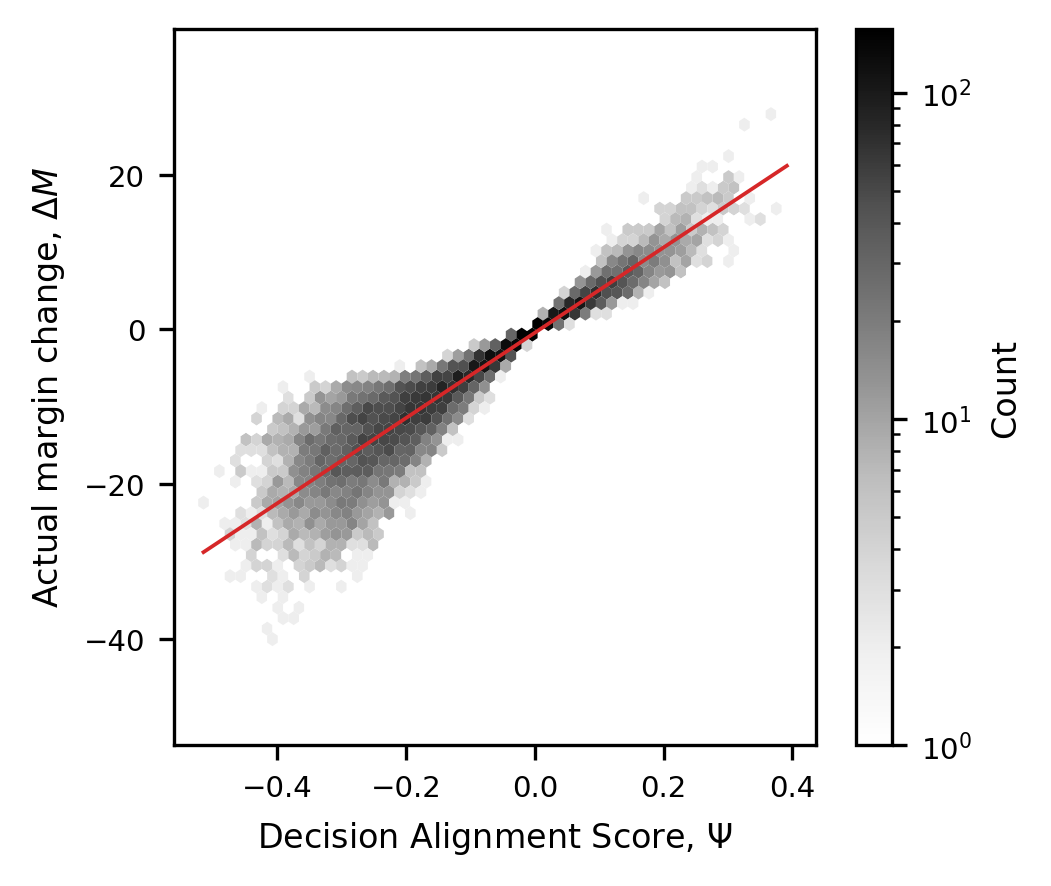}
\par\vspace{-1pt}
{\scriptsize ResNet-50}
\end{minipage}\hfill%
\begin{minipage}[c]{0.192\linewidth}
\centering
\includegraphics[width=\linewidth]{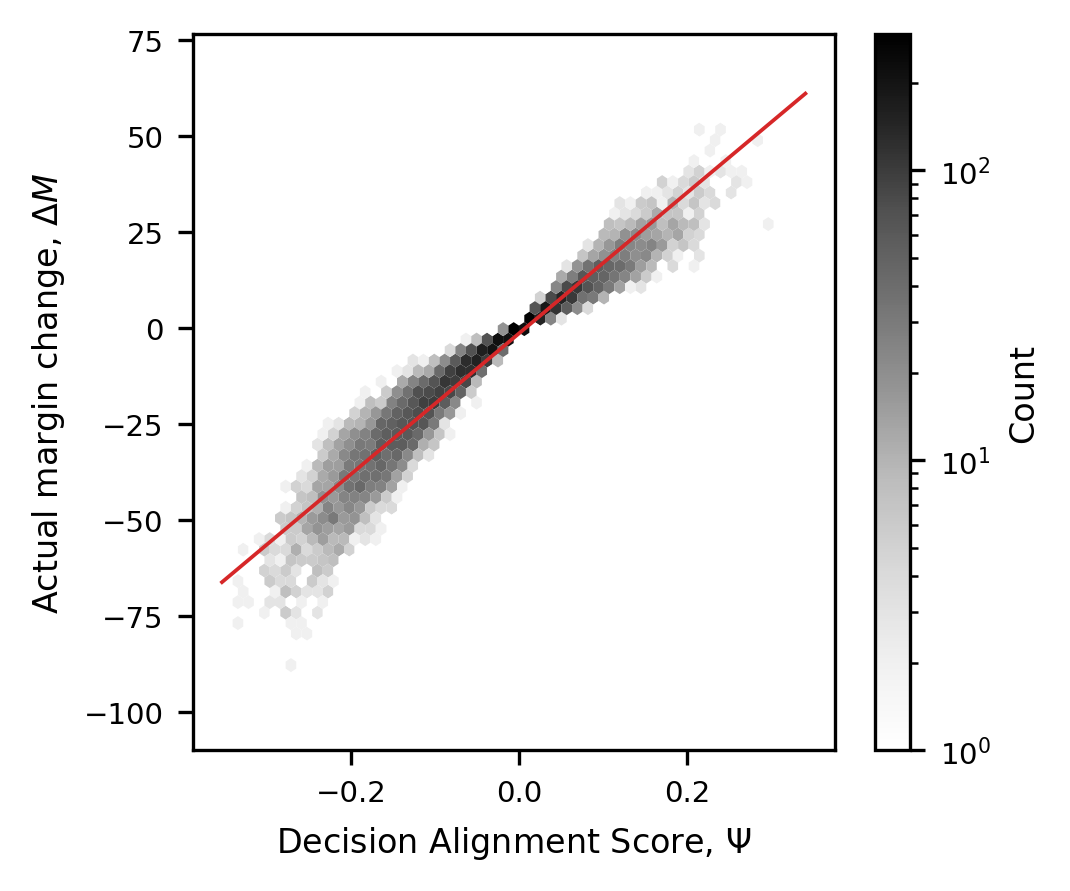}
\par\vspace{-1pt}
{\scriptsize ResNet-56}
\end{minipage}\hfill%
\begin{minipage}[c]{0.192\linewidth}
\centering
\includegraphics[width=\linewidth]{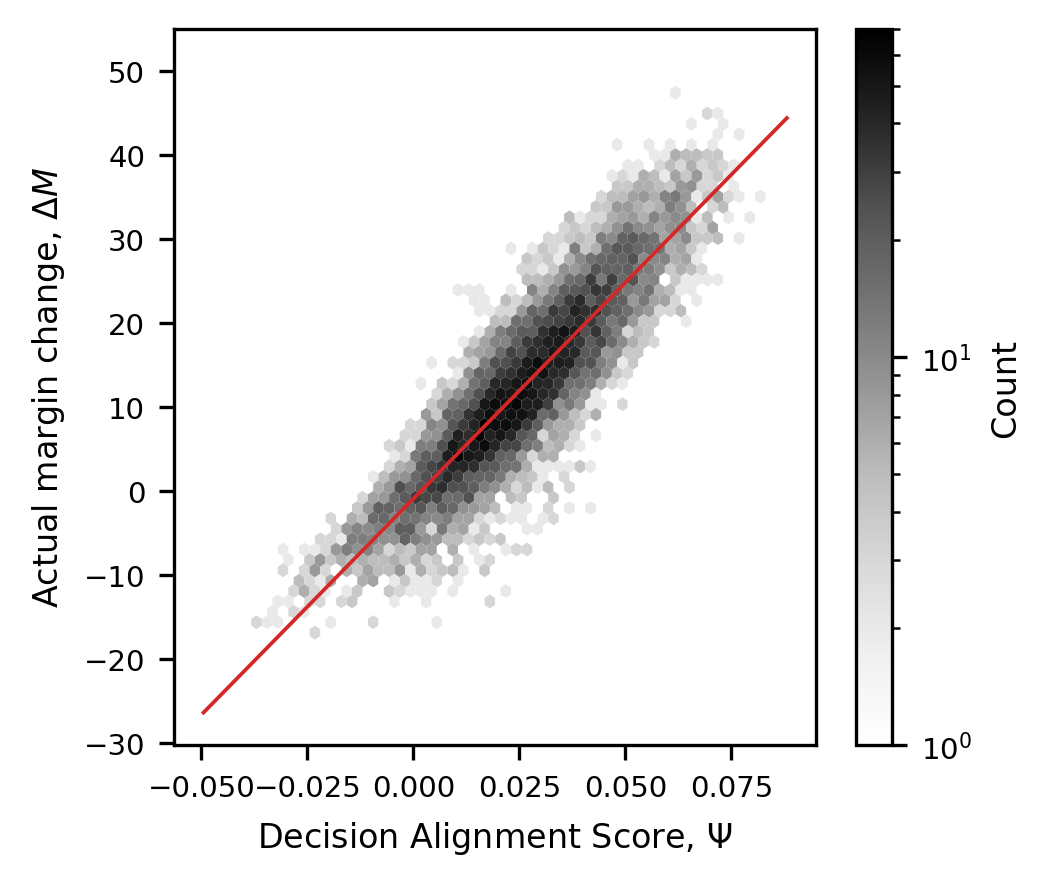}
\par\vspace{-1pt}
{\scriptsize MobileNetV2}
\end{minipage}%
\end{minipage}\par

\vspace{1mm}
\caption{DAS versus actual margin change for the final attention modules in ALL-trained models across five CNN backbones. Results are shown separately for SimAM and CBAM. The red line represents the linear regression fit.}
\label{fig:cifar-alignment-all}

\pairedfigspace

\noindent
\finalStudyRowLabel{SimAM}\hfill%
\begin{minipage}[c]{0.935\textwidth}
\centering
\begin{minipage}[c]{0.205\linewidth}
\centering
\includegraphics[width=\linewidth]{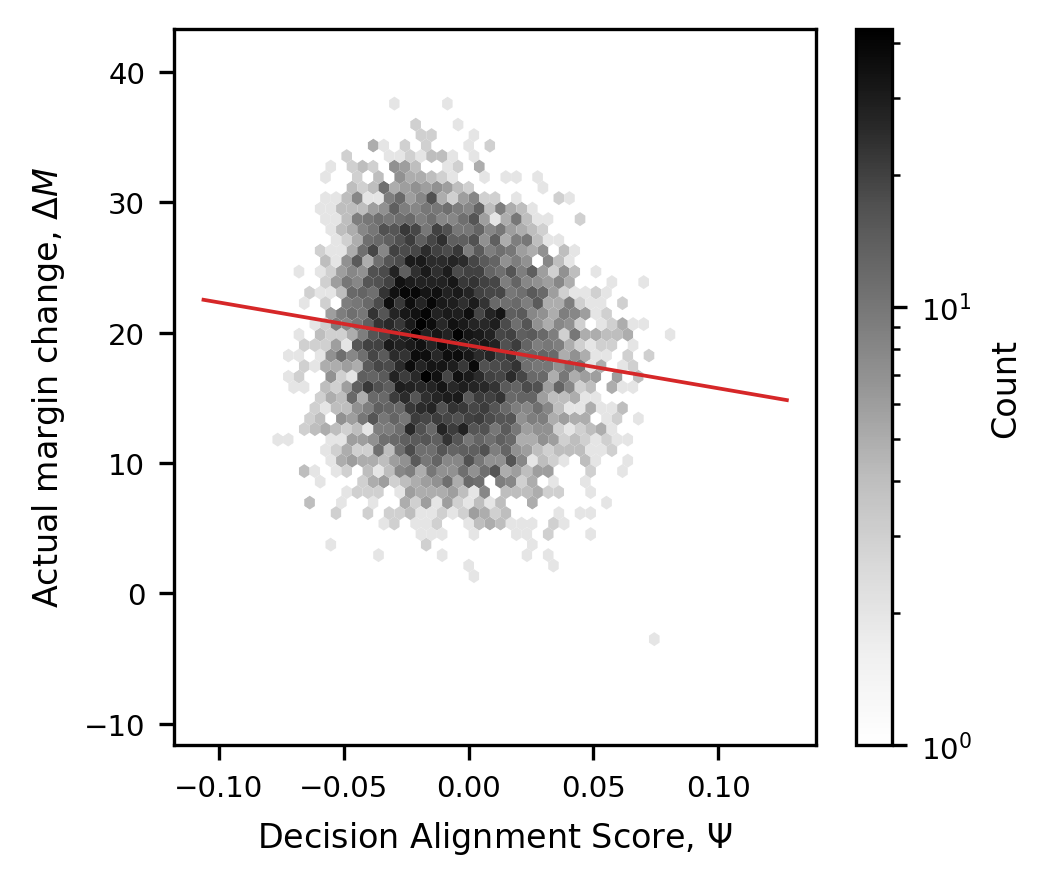}
\par\vspace{-1pt}
{\scriptsize S1}
\end{minipage}\hfill%
\begin{minipage}[c]{0.205\linewidth}
\centering
\includegraphics[width=\linewidth]{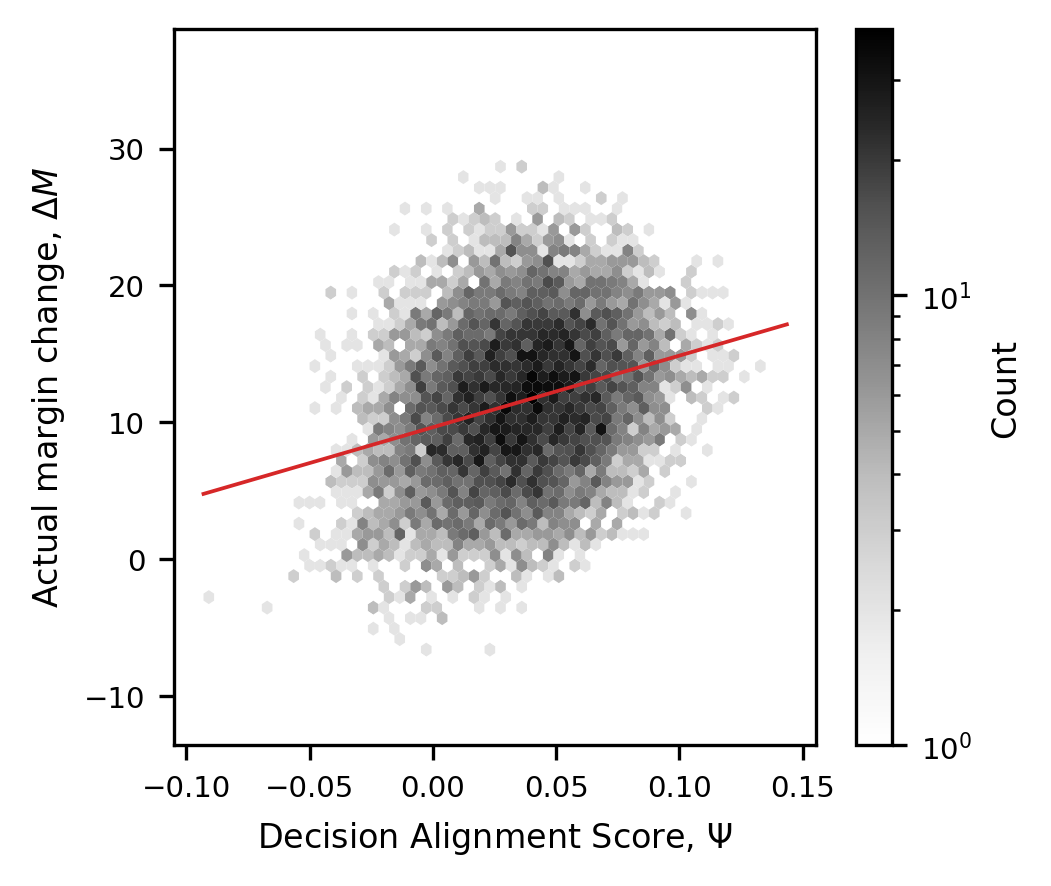}
\par\vspace{-1pt}
{\scriptsize S2}
\end{minipage}\hfill%
\begin{minipage}[c]{0.205\linewidth}
\centering
\includegraphics[width=\linewidth]{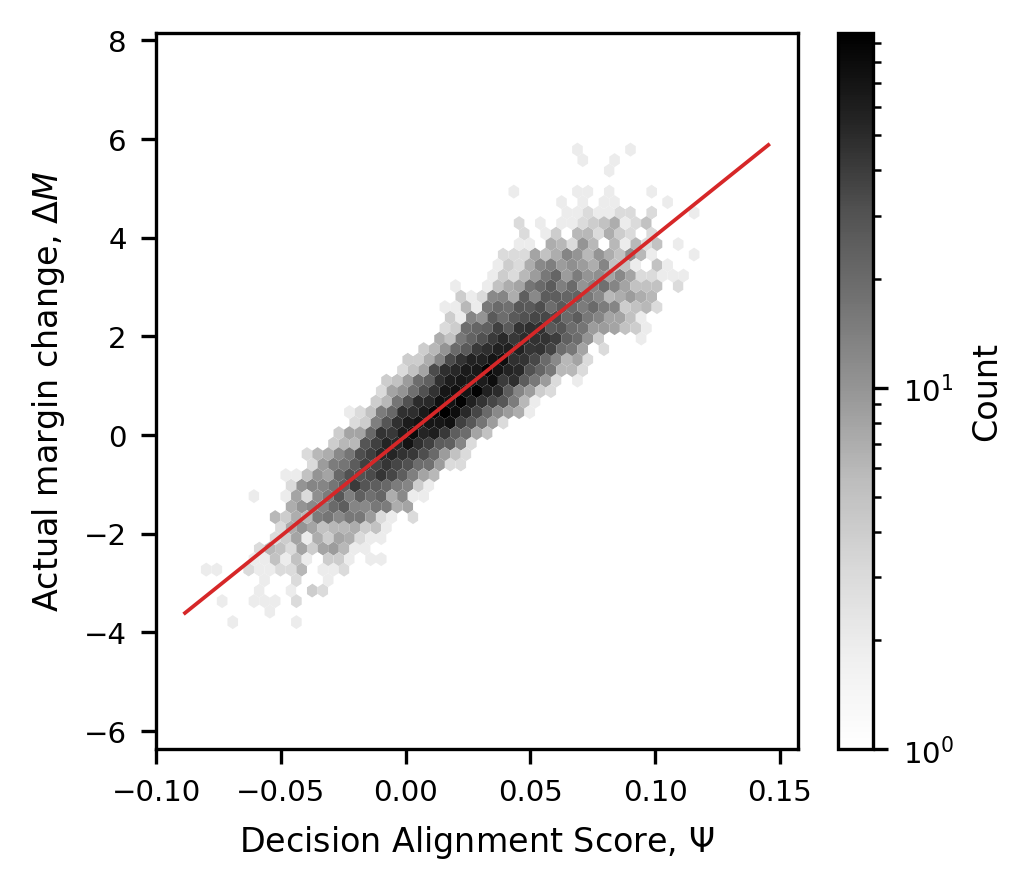}
\par\vspace{-1pt}
{\scriptsize S3}
\end{minipage}\hfill%
\begin{minipage}[c]{0.205\linewidth}
\centering
\includegraphics[width=\linewidth]{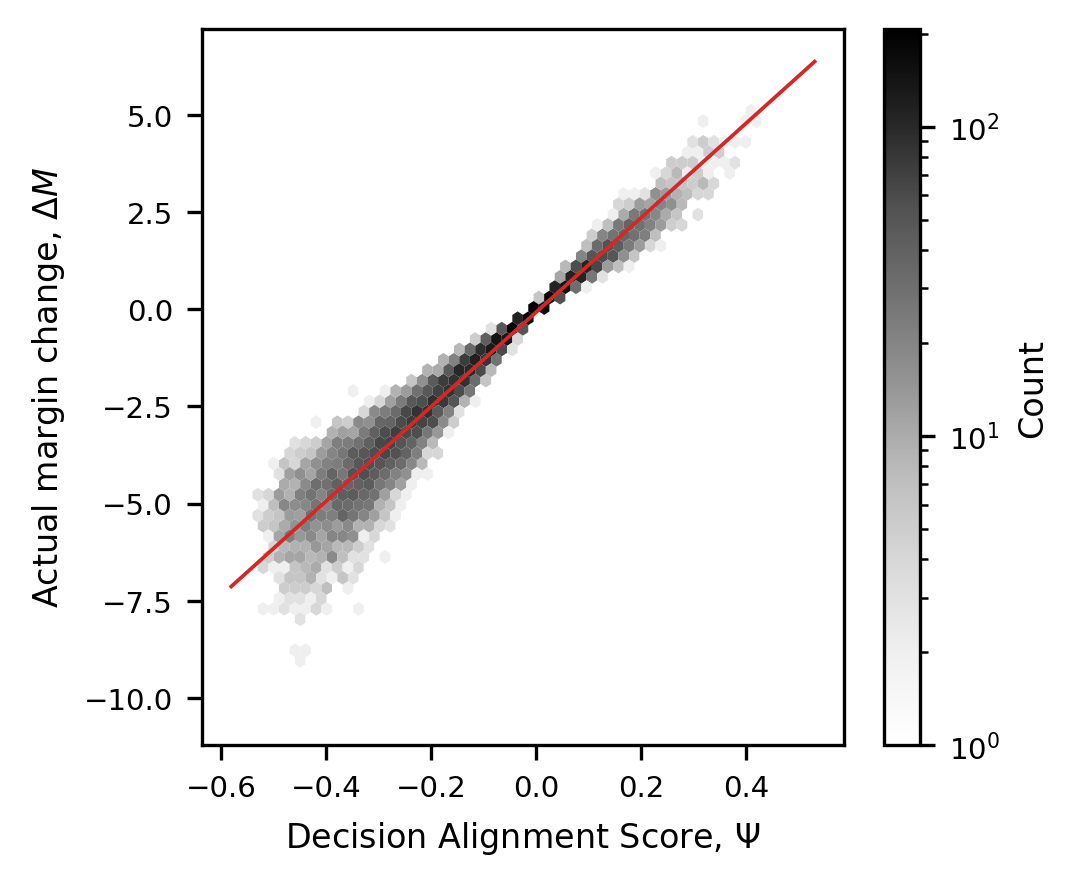}
\par\vspace{-1pt}
{\scriptsize S4}
\end{minipage}%
\end{minipage}\par

\vspace{2mm}

\noindent
\finalStudyRowLabel{CBAM}\hfill%
\begin{minipage}[c]{0.935\textwidth}
\centering
\begin{minipage}[c]{0.205\linewidth}
\centering
\includegraphics[width=\linewidth]{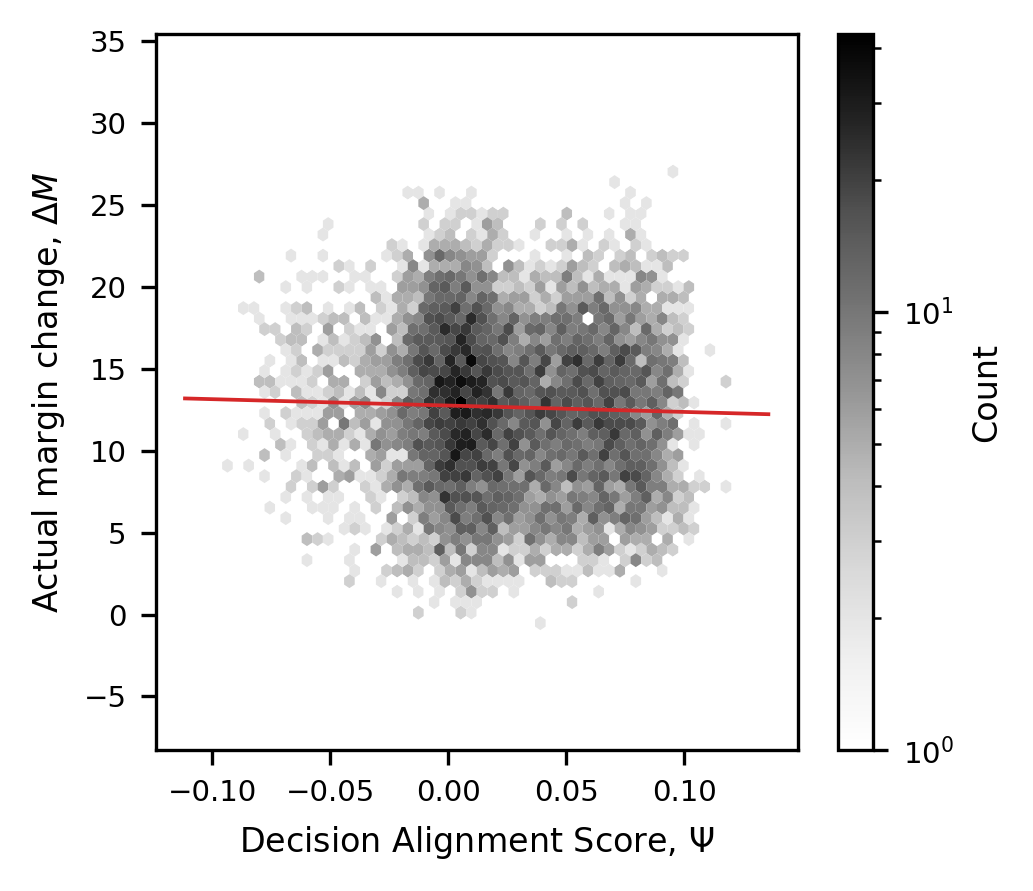}
\par\vspace{-1pt}
{\scriptsize S1}
\end{minipage}\hfill%
\begin{minipage}[c]{0.205\linewidth}
\centering
\includegraphics[width=\linewidth]{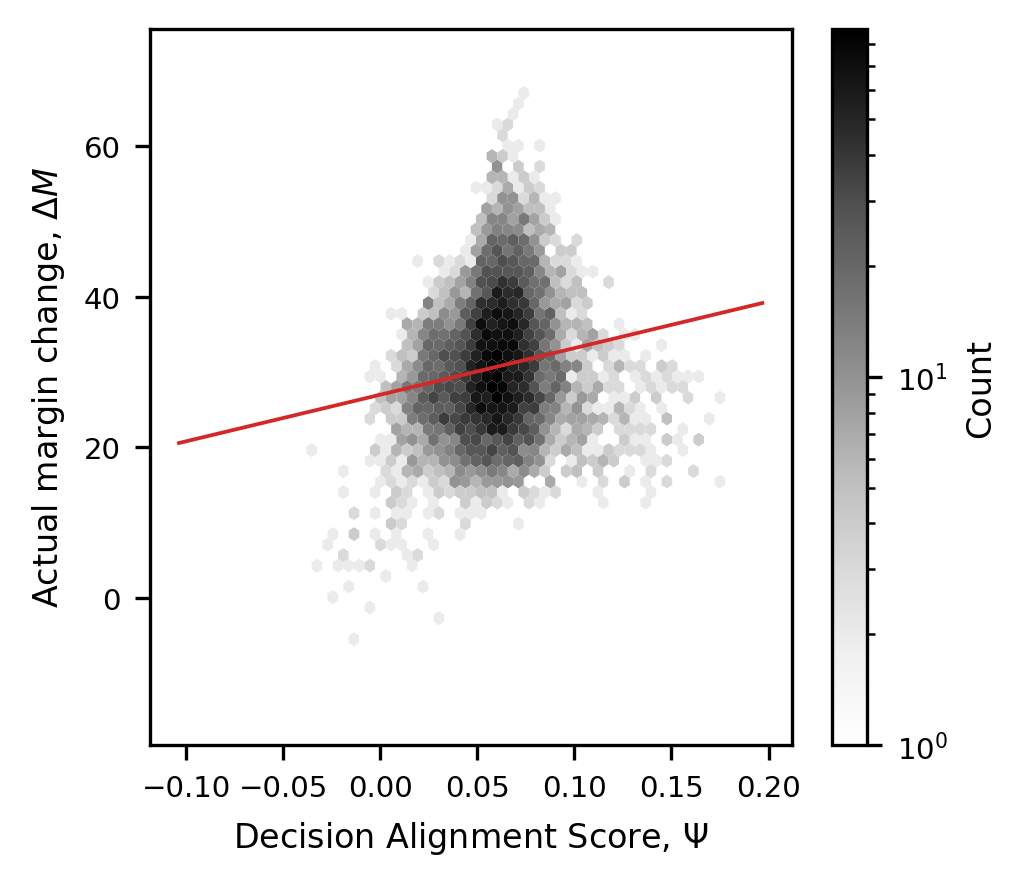}
\par\vspace{-1pt}
{\scriptsize S2}
\end{minipage}\hfill%
\begin{minipage}[c]{0.205\linewidth}
\centering
\includegraphics[width=\linewidth]{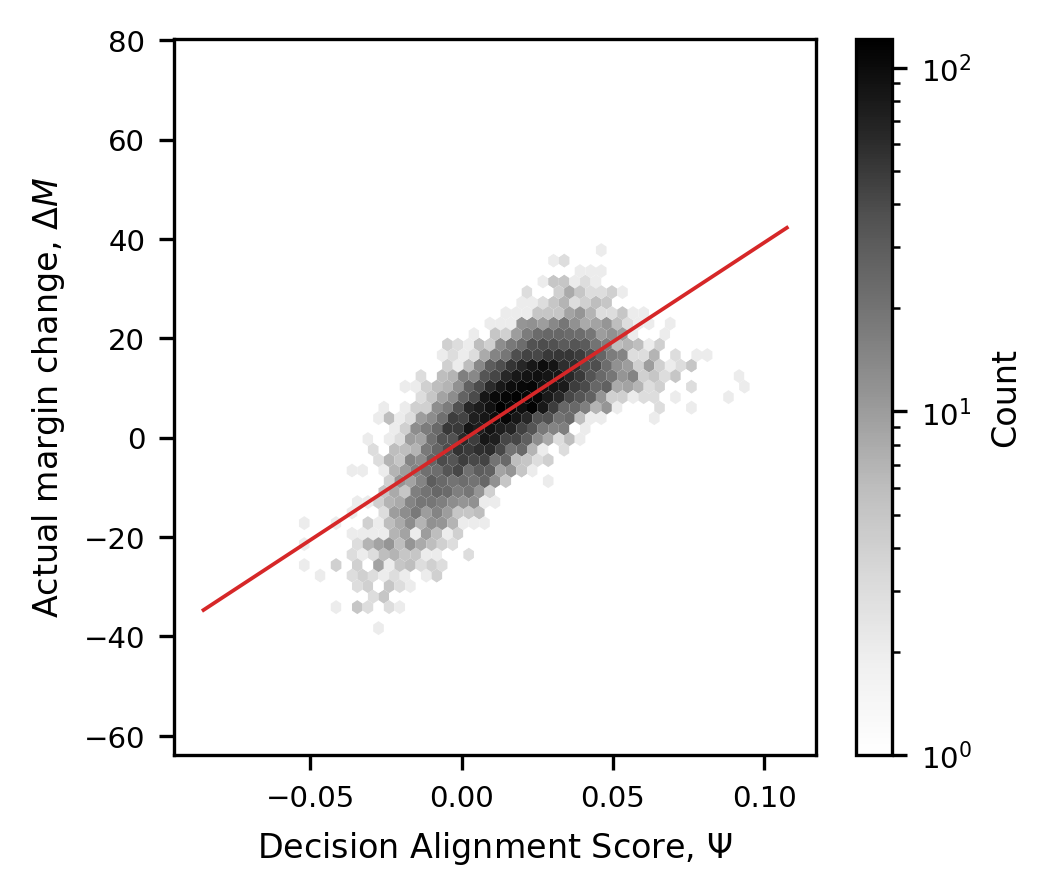}
\par\vspace{-1pt}
{\scriptsize S3}
\end{minipage}\hfill%
\begin{minipage}[c]{0.205\linewidth}
\centering
\includegraphics[width=\linewidth]{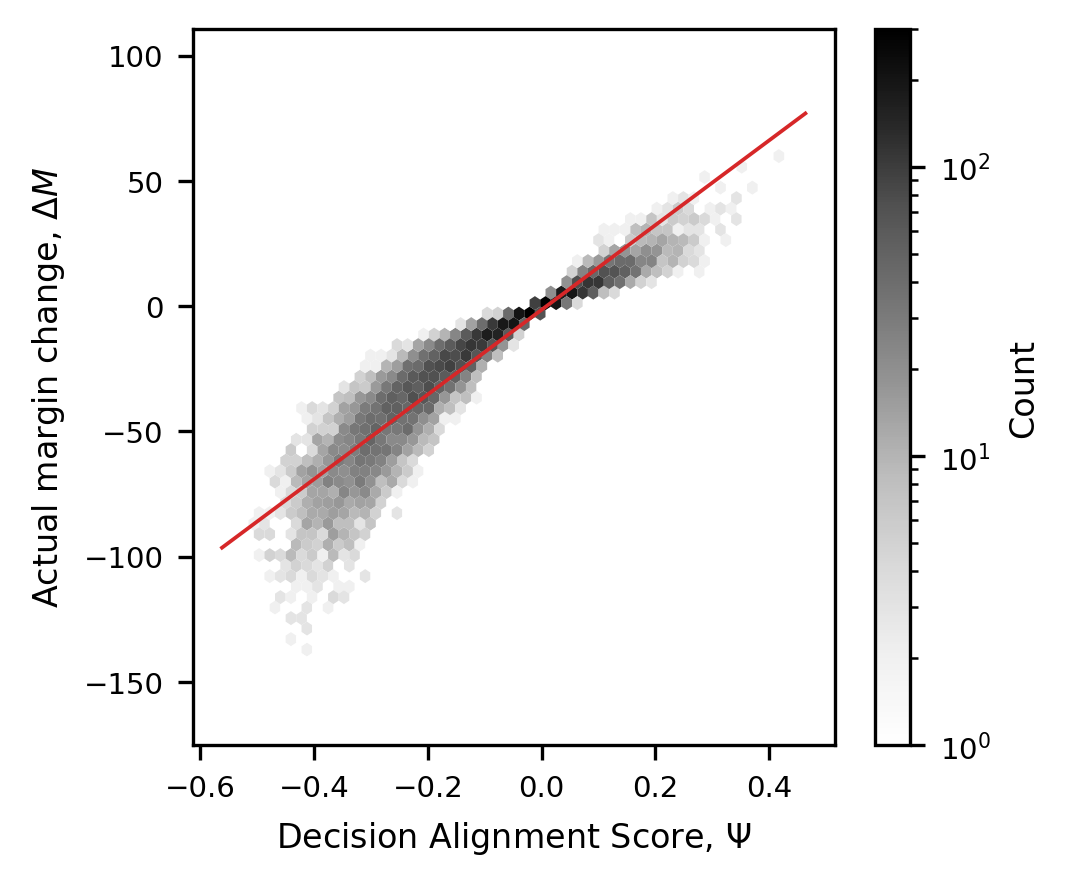}
\par\vspace{-1pt}
{\scriptsize S4}
\end{minipage}%
\end{minipage}\par

\vspace{1mm}
\caption{DAS versus actual margin change for independently trained single-attention module models (S1--S4) using SimAM and CBAM. The red line represents the linear regression fit.}
\label{fig:cifar-alignment-single}

\end{figure*}

\begin{figure*}[]
\centering

\noindent
\finalStudyRowLabel{SimAM}\hfill%
\begin{minipage}[c]{0.935\textwidth}
\centering
\begin{minipage}[c]{0.192\linewidth}
\centering
\includegraphics[width=\linewidth]{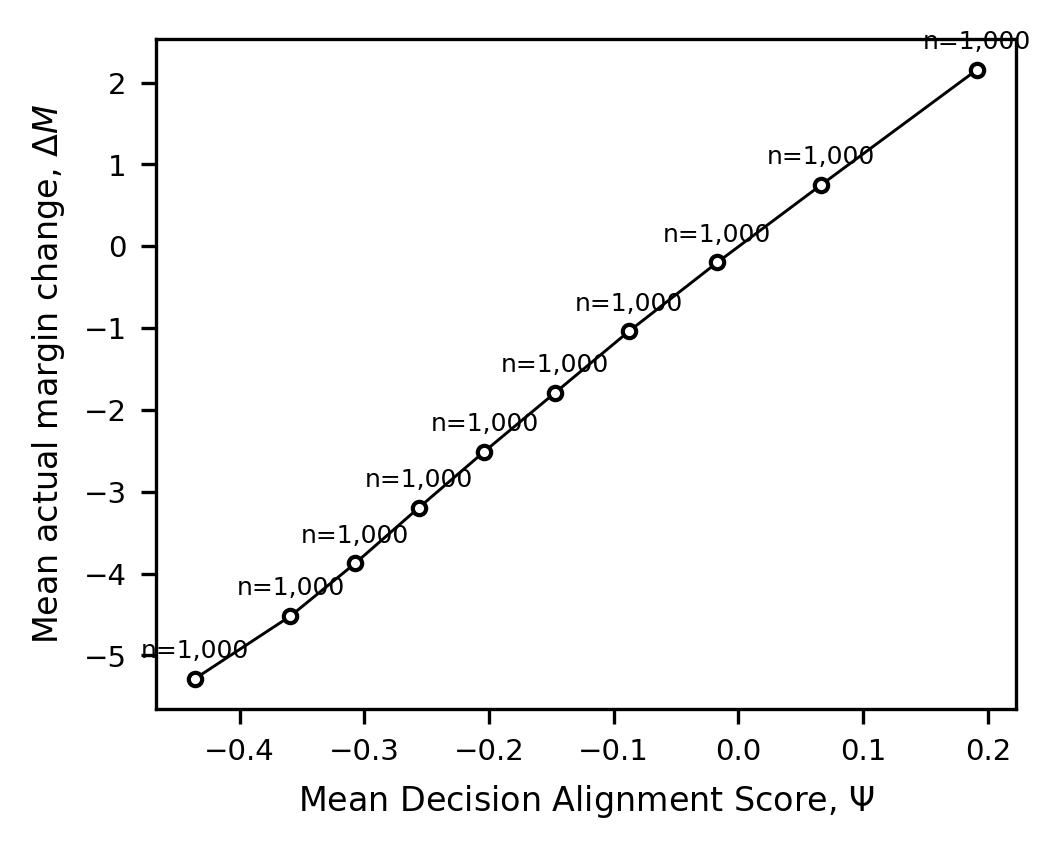}
\par\vspace{-1pt}
{\scriptsize ResNet-18}
\end{minipage}\hfill%
\begin{minipage}[c]{0.192\linewidth}
\centering
\includegraphics[width=\linewidth]{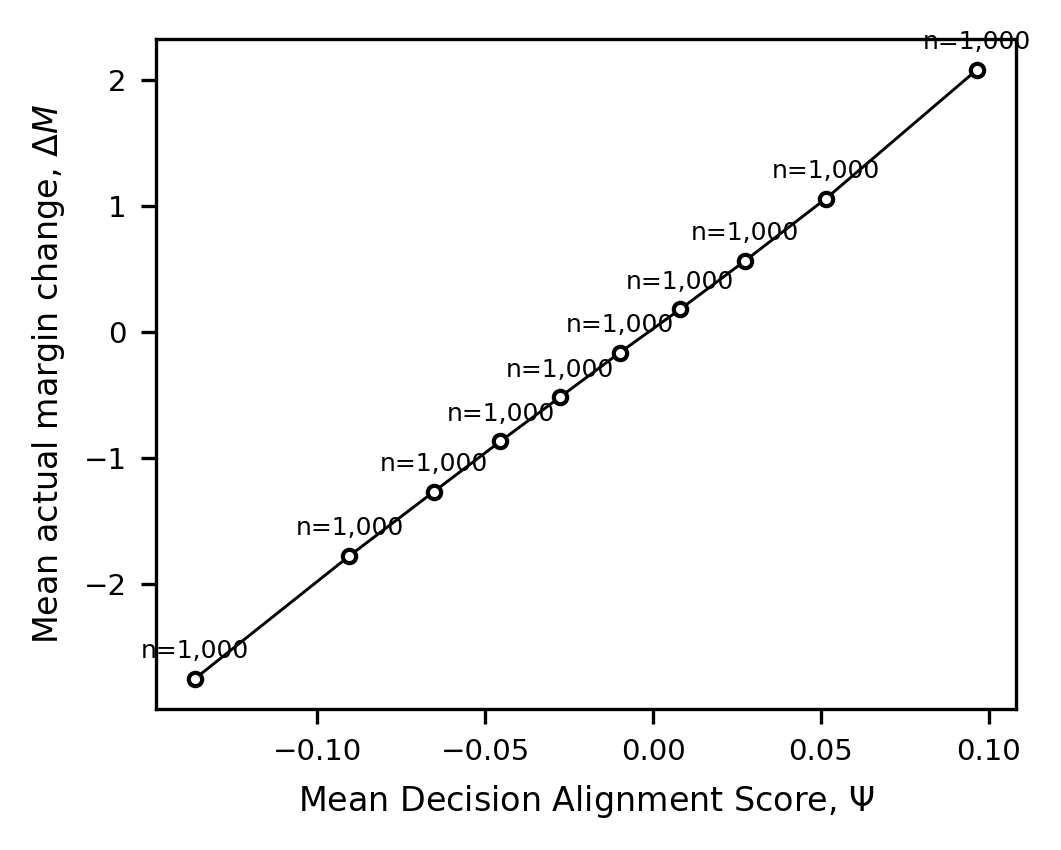}
\par\vspace{-1pt}
{\scriptsize ResNet-20}
\end{minipage}\hfill%
\begin{minipage}[c]{0.192\linewidth}
\centering
\includegraphics[width=\linewidth]{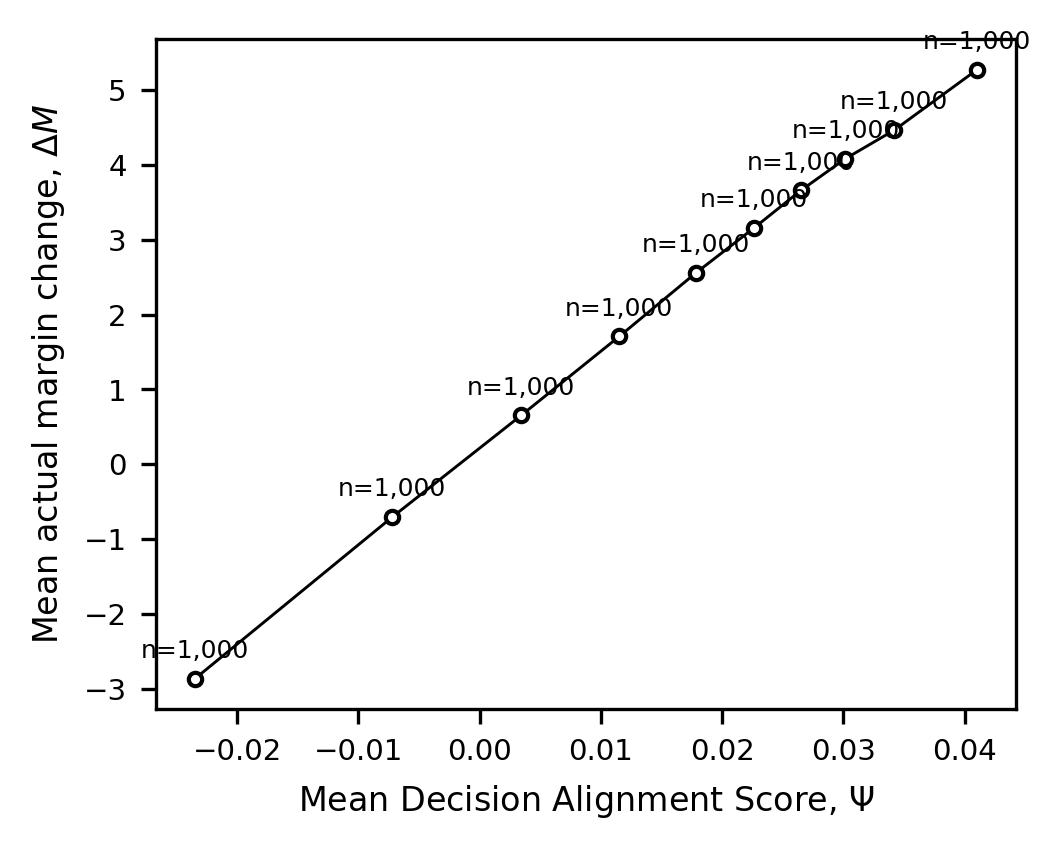}
\par\vspace{-1pt}
{\scriptsize ResNet-50}
\end{minipage}\hfill%
\begin{minipage}[c]{0.192\linewidth}
\centering
\includegraphics[width=\linewidth]{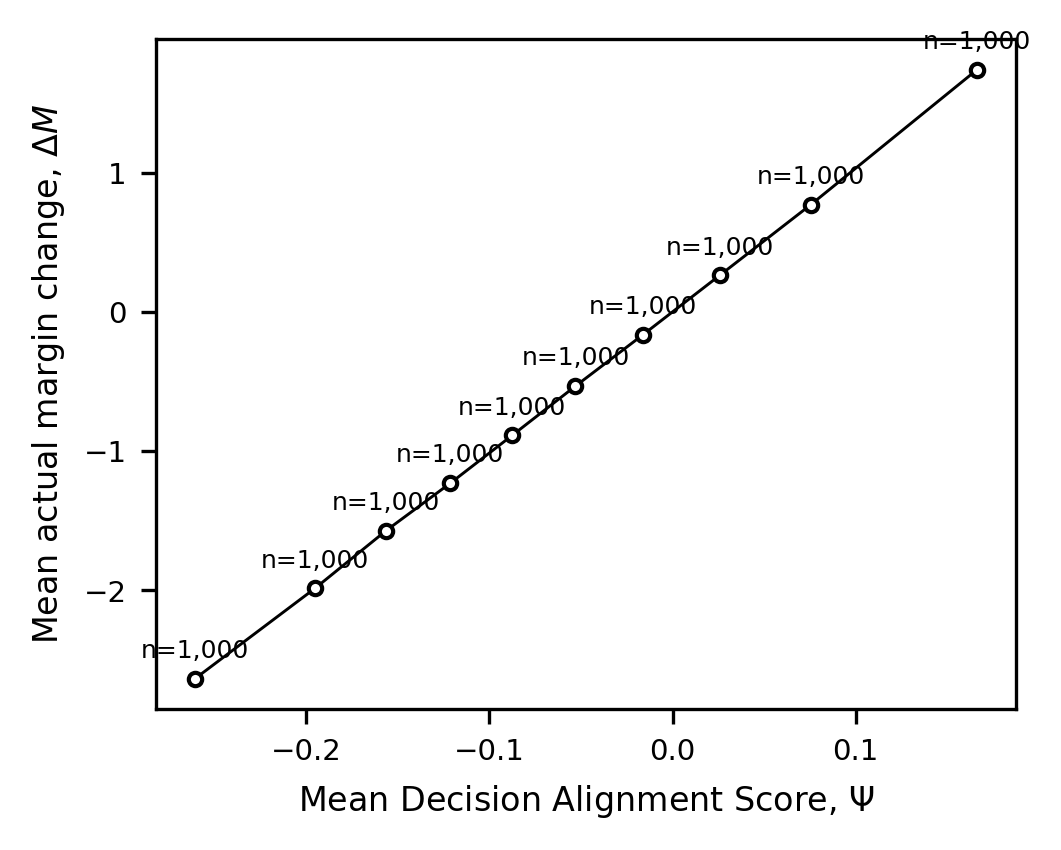}
\par\vspace{-1pt}
{\scriptsize ResNet-56}
\end{minipage}\hfill%
\begin{minipage}[c]{0.192\linewidth}
\centering
\includegraphics[width=\linewidth]{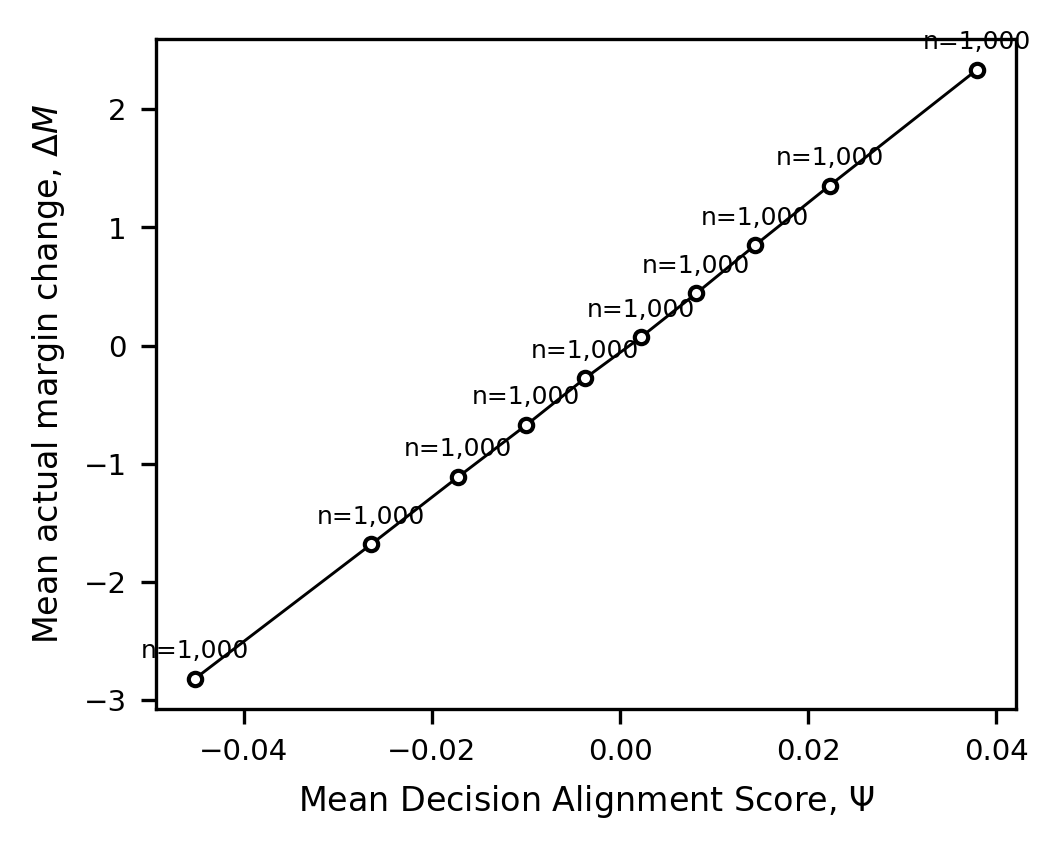}
\par\vspace{-1pt}
{\scriptsize MobileNetV2}
\end{minipage}%
\end{minipage}\par

\vspace{2mm}

\noindent
\finalStudyRowLabel{CBAM}\hfill%
\begin{minipage}[c]{0.935\textwidth}
\centering
\begin{minipage}[c]{0.192\linewidth}
\centering
\includegraphics[width=\linewidth]{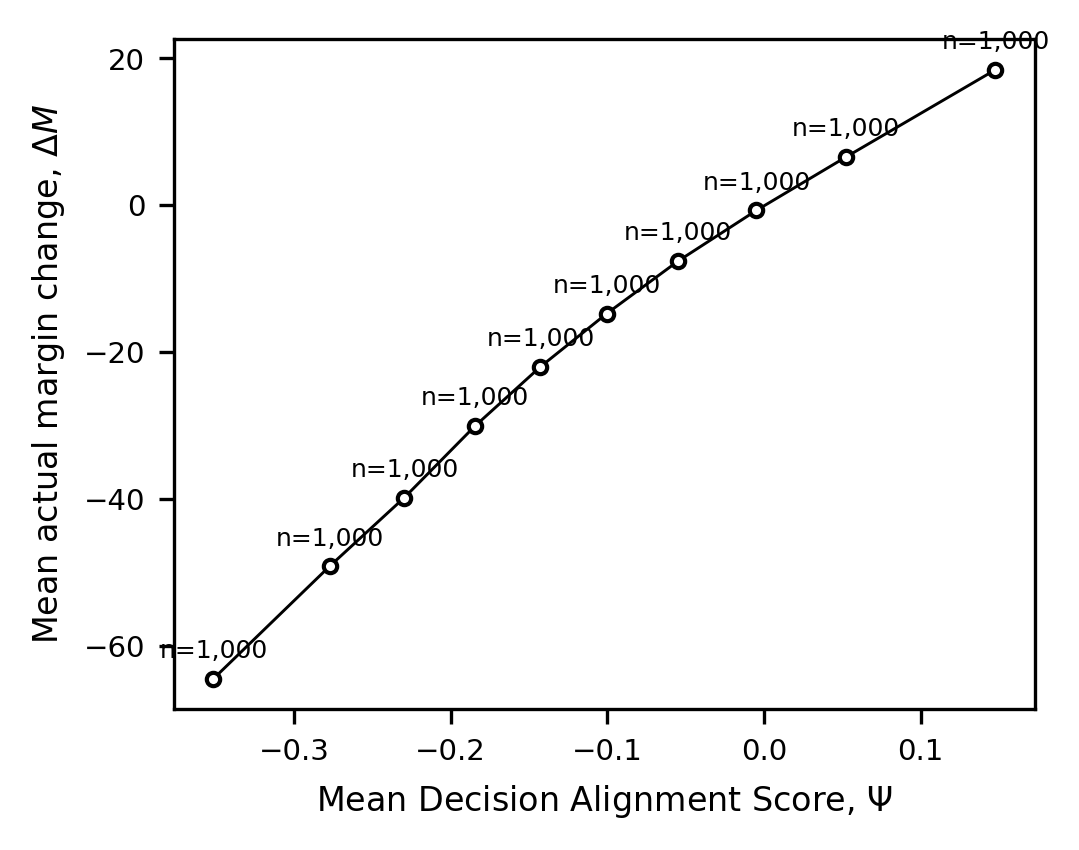}
\par\vspace{-1pt}
{\scriptsize ResNet-18}
\end{minipage}\hfill%
\begin{minipage}[c]{0.192\linewidth}
\centering
\includegraphics[width=\linewidth]{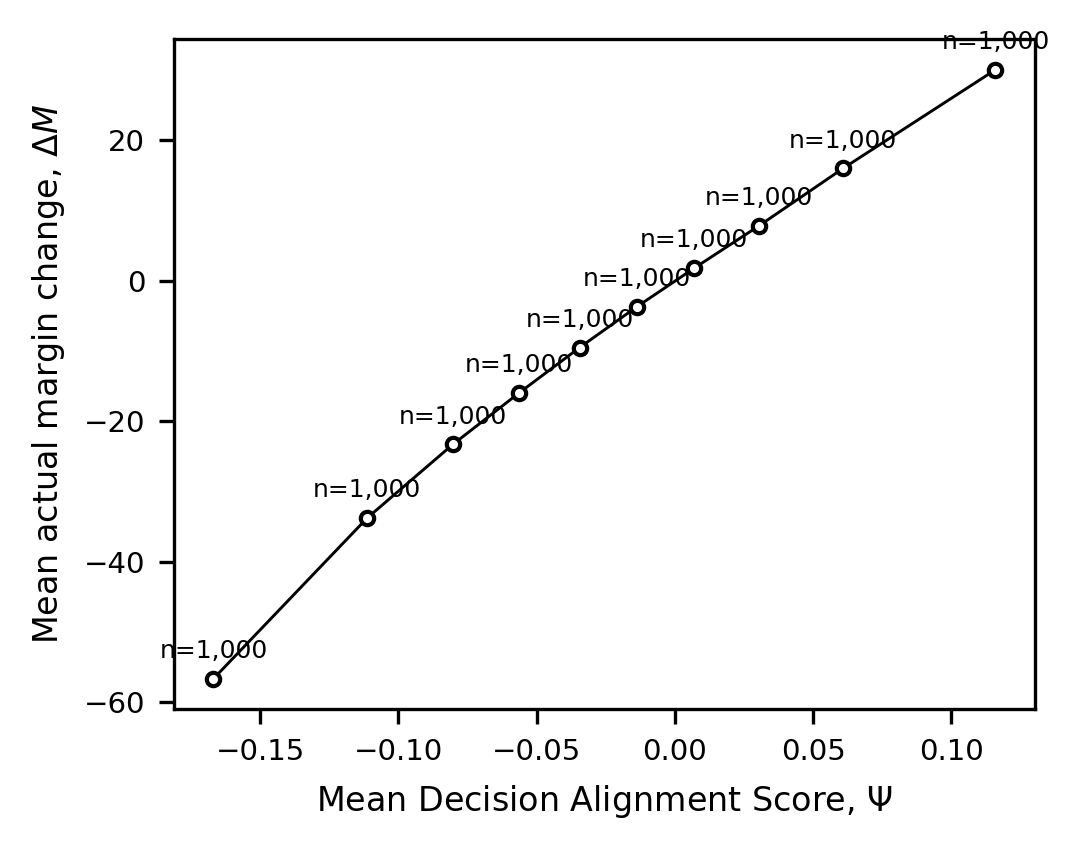}
\par\vspace{-1pt}
{\scriptsize ResNet-20}
\end{minipage}\hfill%
\begin{minipage}[c]{0.192\linewidth}
\centering
\includegraphics[width=\linewidth]{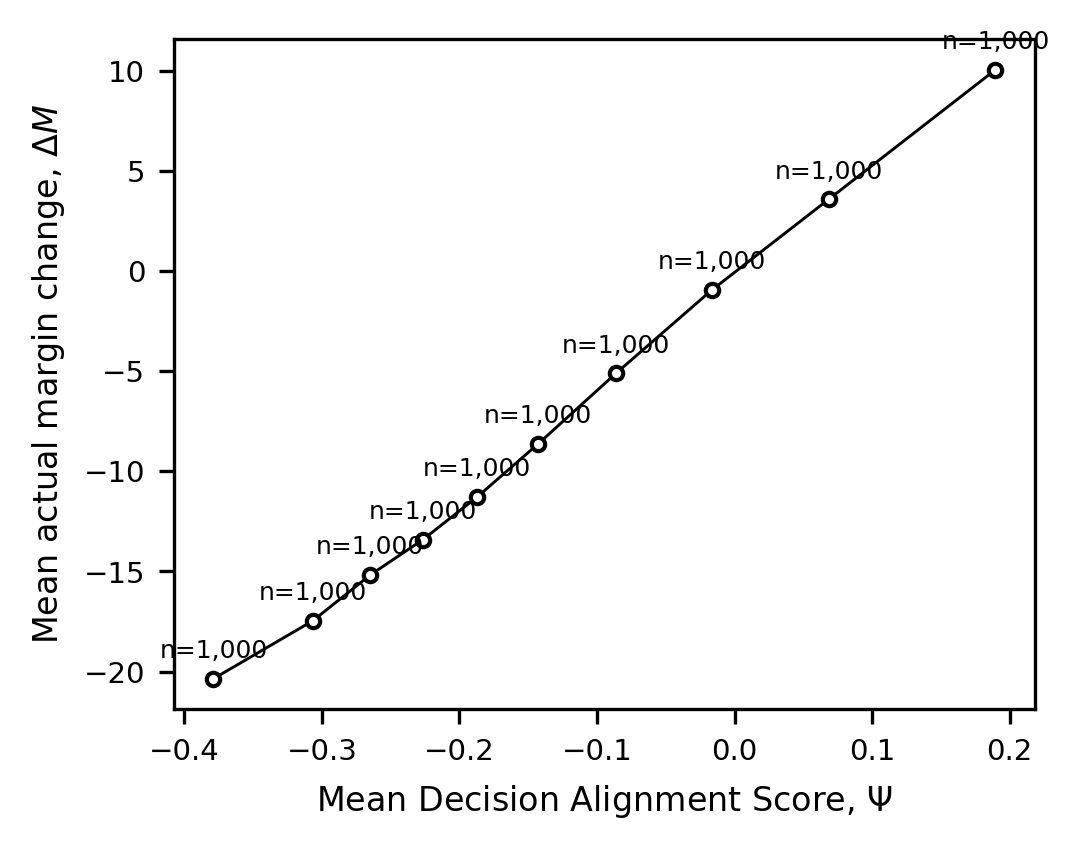}
\par\vspace{-1pt}
{\scriptsize ResNet-50}
\end{minipage}\hfill%
\begin{minipage}[c]{0.192\linewidth}
\centering
\includegraphics[width=\linewidth]{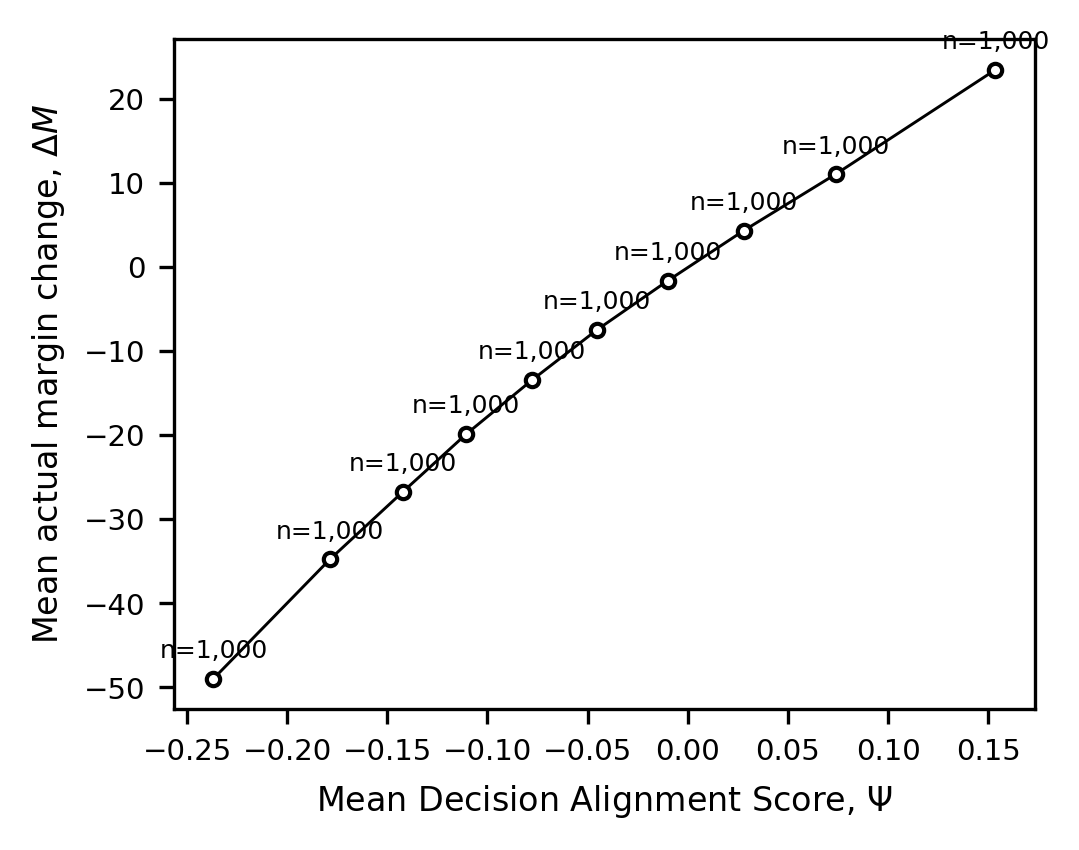}
\par\vspace{-1pt}
{\scriptsize ResNet-56}
\end{minipage}\hfill%
\begin{minipage}[c]{0.192\linewidth}
\centering
\includegraphics[width=\linewidth]{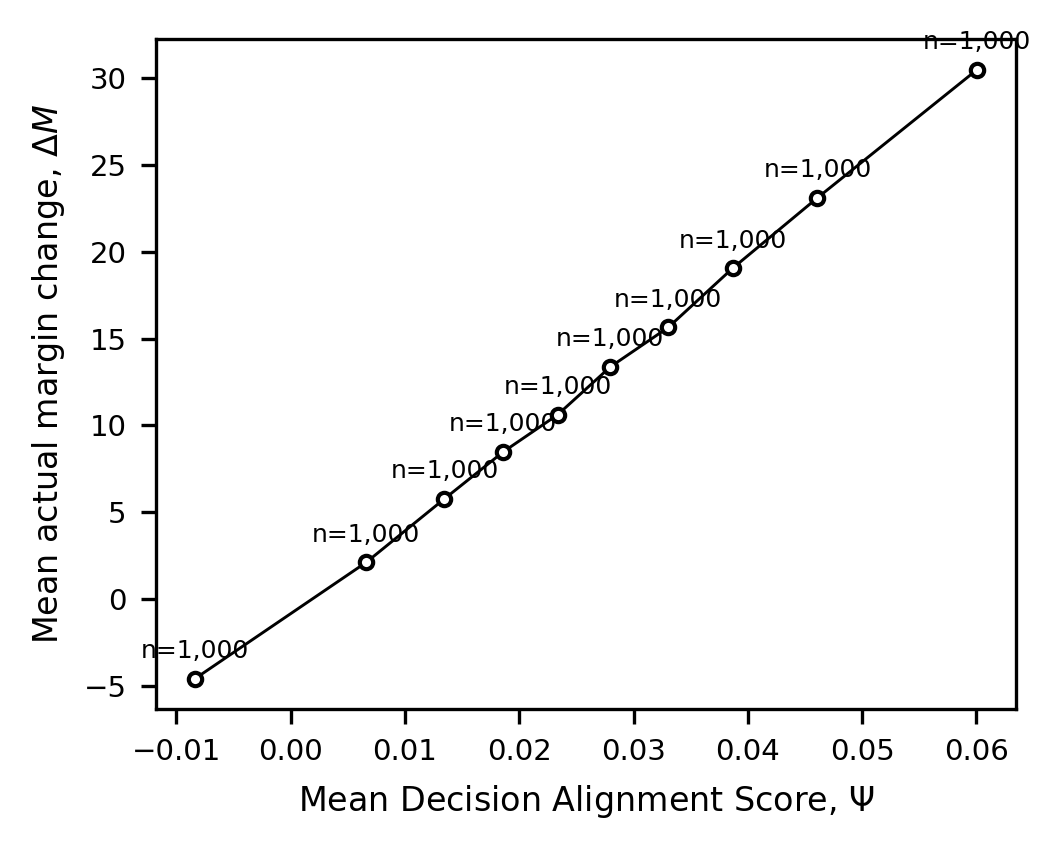}
\par\vspace{-1pt}
{\scriptsize MobileNetV2}
\end{minipage}%
\end{minipage}\par

\vspace{1mm}
\caption{Equal-count DAS-group analysis for the final attention modules in ALL-trained models. Each point represents the mean DAS and actual margin change within a group; annotations indicate group sizes. Best viewed with zoom factor for clarity.}
\label{fig:cifar-deciles-all}

\pairedfigspace

\noindent
\finalStudyRowLabel{SimAM}\hfill%
\begin{minipage}[c]{0.935\textwidth}
\centering
\begin{minipage}[c]{0.205\linewidth}
\centering
\includegraphics[width=\linewidth]{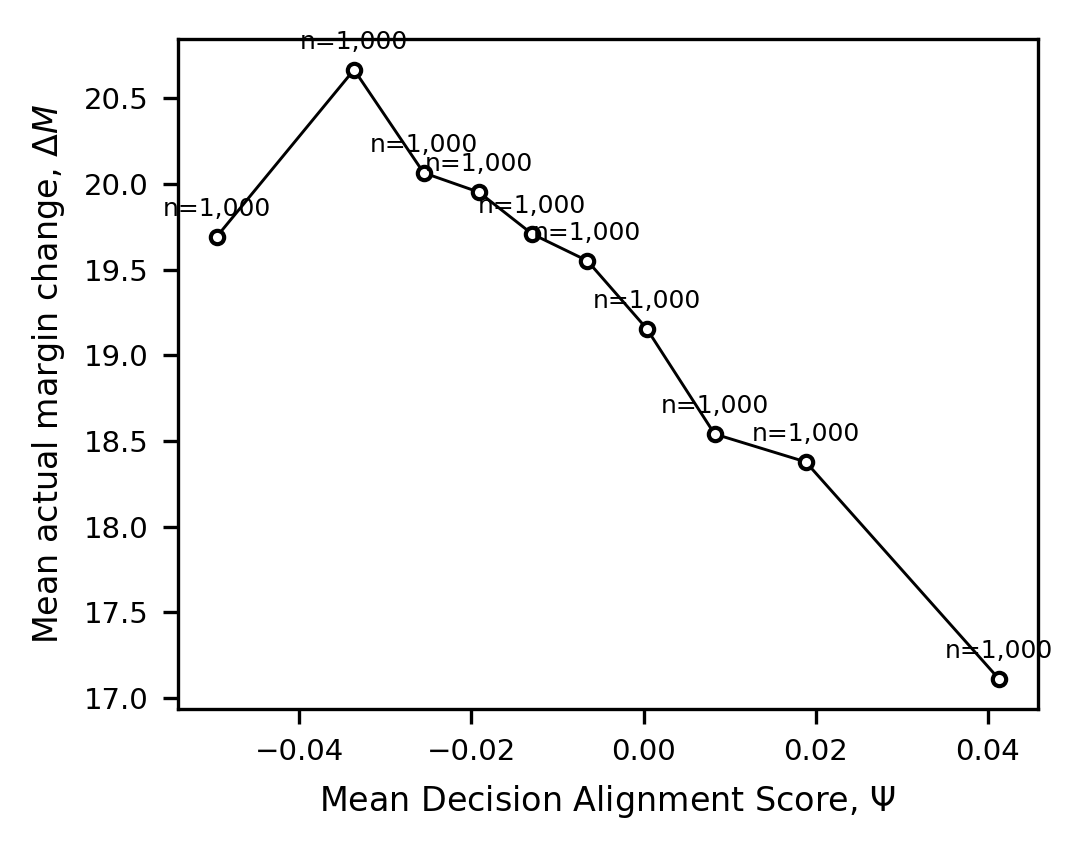}
\par\vspace{-1pt}
{\scriptsize S1}
\end{minipage}\hfill%
\begin{minipage}[c]{0.205\linewidth}
\centering
\includegraphics[width=\linewidth]{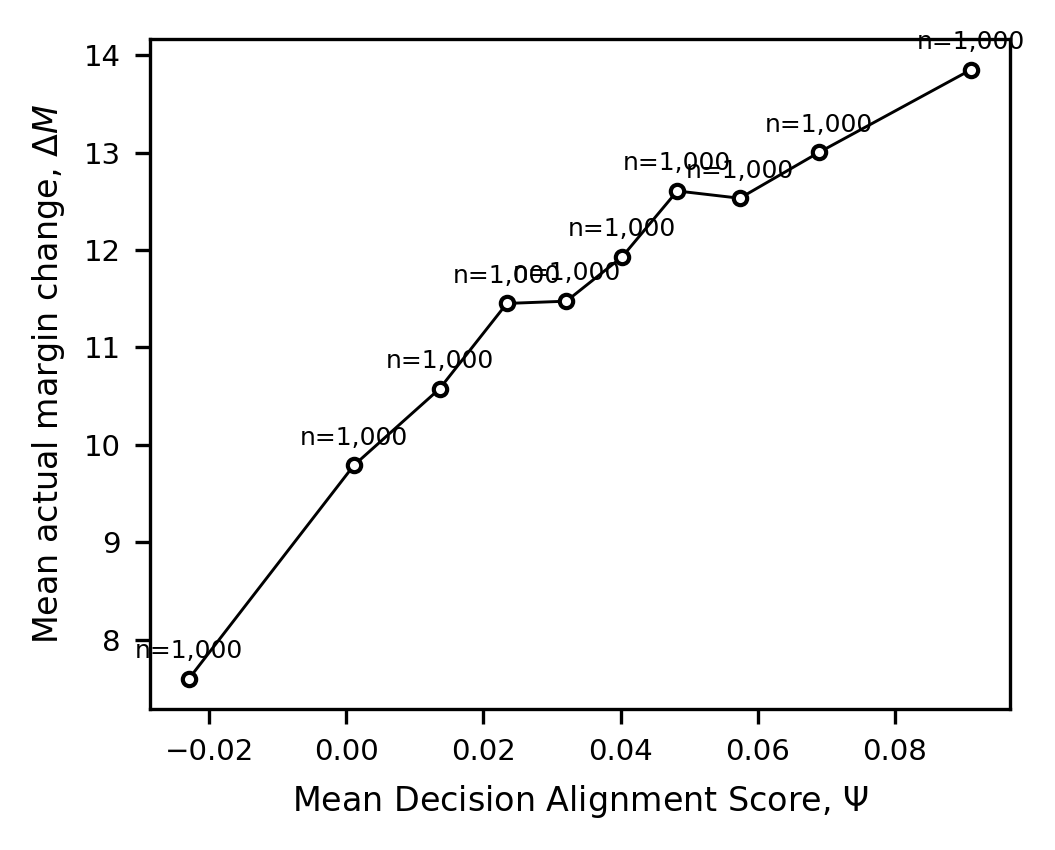}
\par\vspace{-1pt}
{\scriptsize S2}
\end{minipage}\hfill%
\begin{minipage}[c]{0.205\linewidth}
\centering
\includegraphics[width=\linewidth]{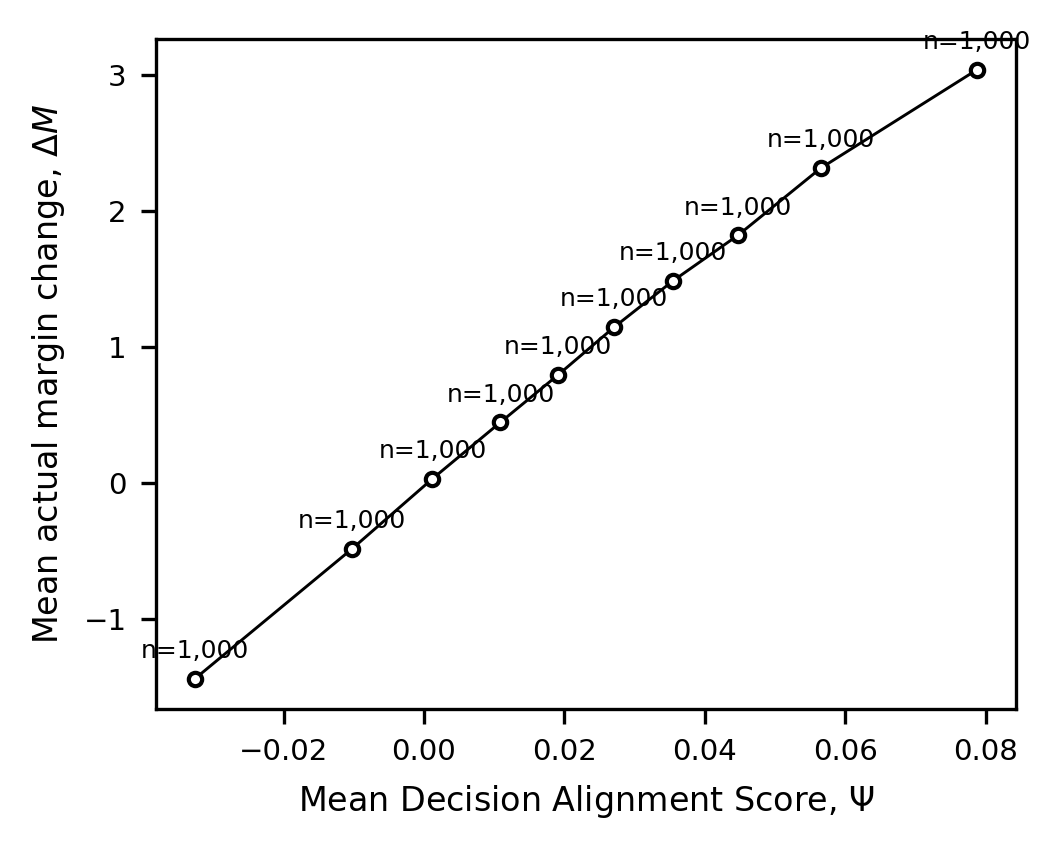}
\par\vspace{-1pt}
{\scriptsize S3}
\end{minipage}\hfill%
\begin{minipage}[c]{0.205\linewidth}
\centering
\includegraphics[width=\linewidth]{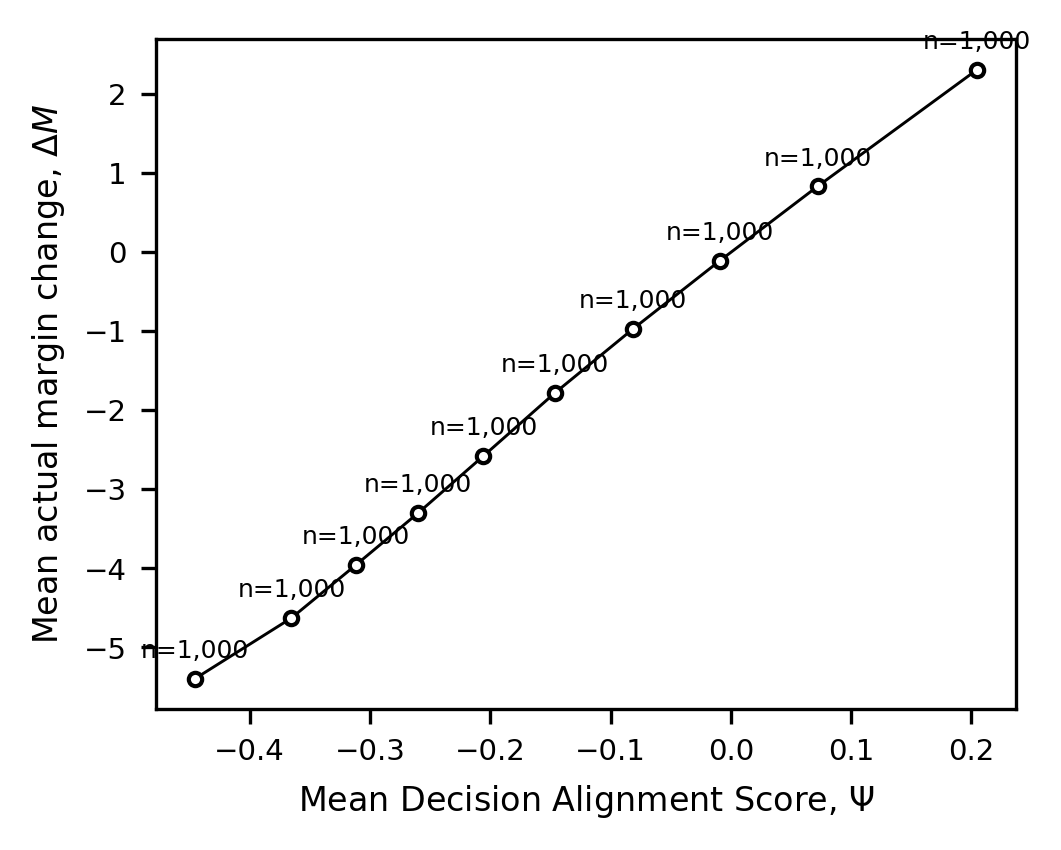}
\par\vspace{-1pt}
{\scriptsize S4}
\end{minipage}%
\end{minipage}\par

\vspace{2mm}

\noindent
\finalStudyRowLabel{CBAM}\hfill%
\begin{minipage}[c]{0.935\textwidth}
\centering
\begin{minipage}[c]{0.205\linewidth}
\centering
\includegraphics[width=\linewidth]{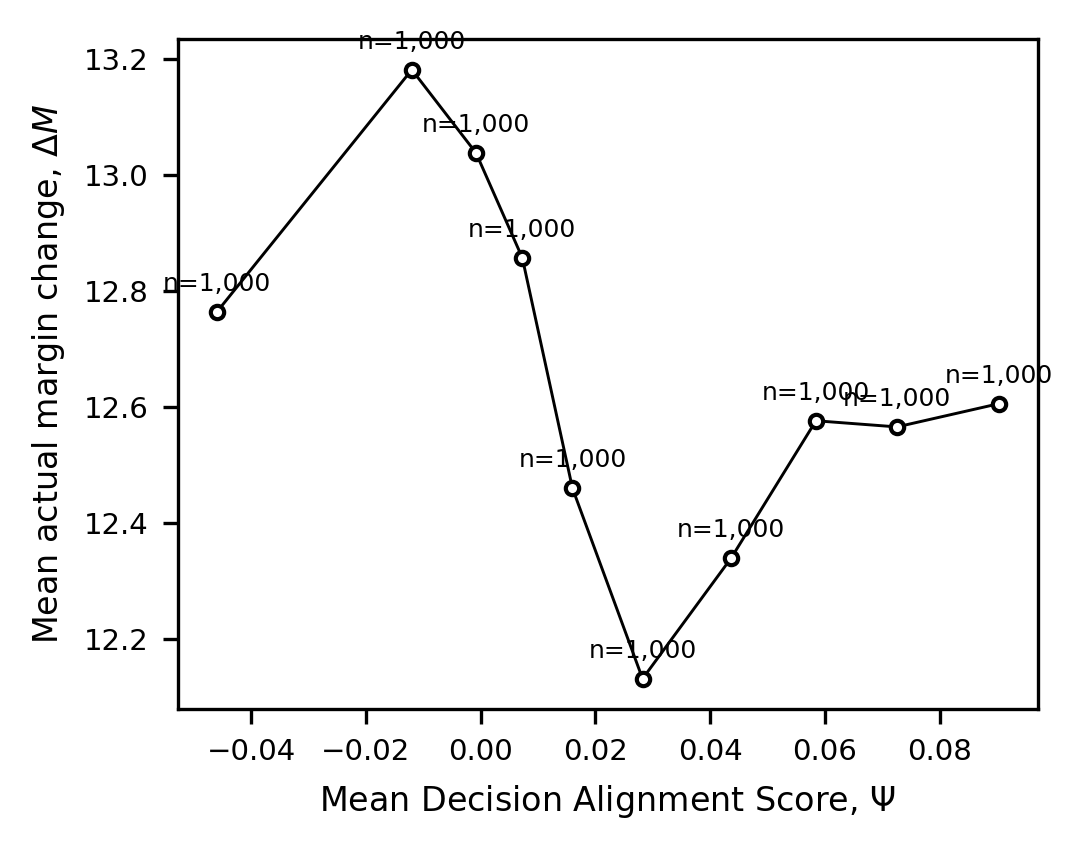}
\par\vspace{-1pt}
{\scriptsize S1}
\end{minipage}\hfill%
\begin{minipage}[c]{0.205\linewidth}
\centering
\includegraphics[width=\linewidth]{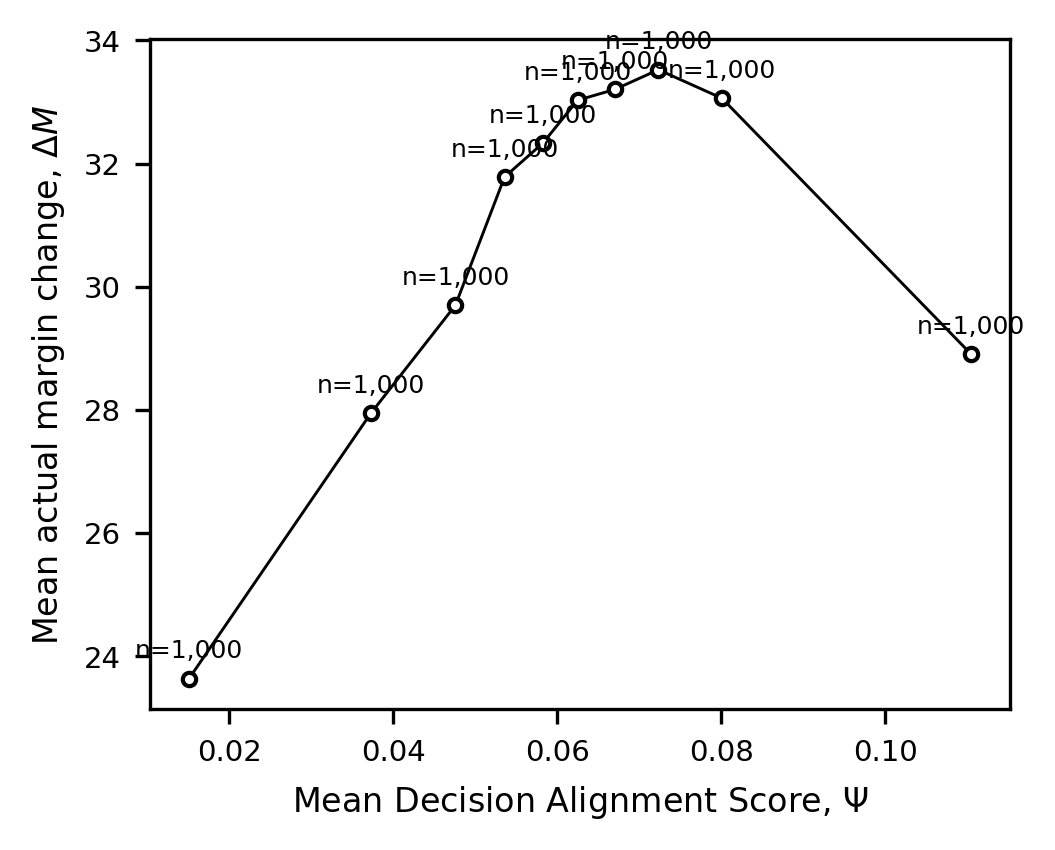}
\par\vspace{-1pt}
{\scriptsize S2}
\end{minipage}\hfill%
\begin{minipage}[c]{0.205\linewidth}
\centering
\includegraphics[width=\linewidth]{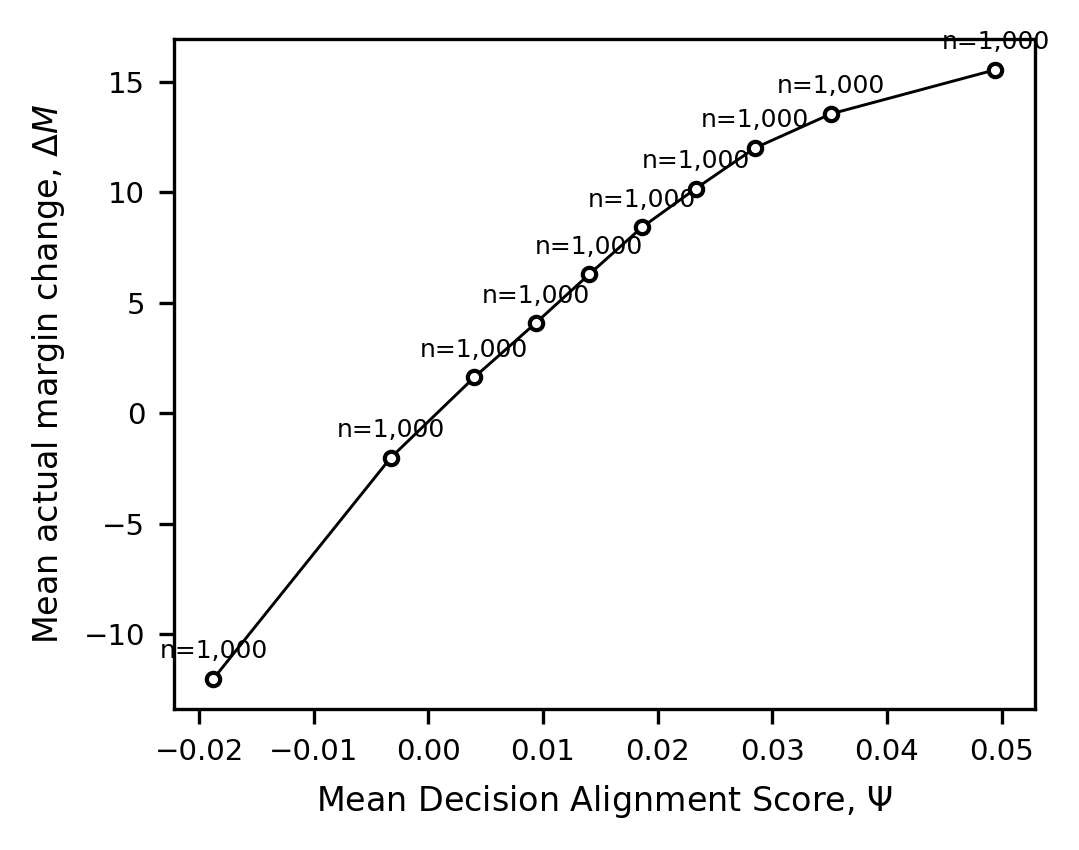}
\par\vspace{-1pt}
{\scriptsize S3}
\end{minipage}\hfill%
\begin{minipage}[c]{0.205\linewidth}
\centering
\includegraphics[width=\linewidth]{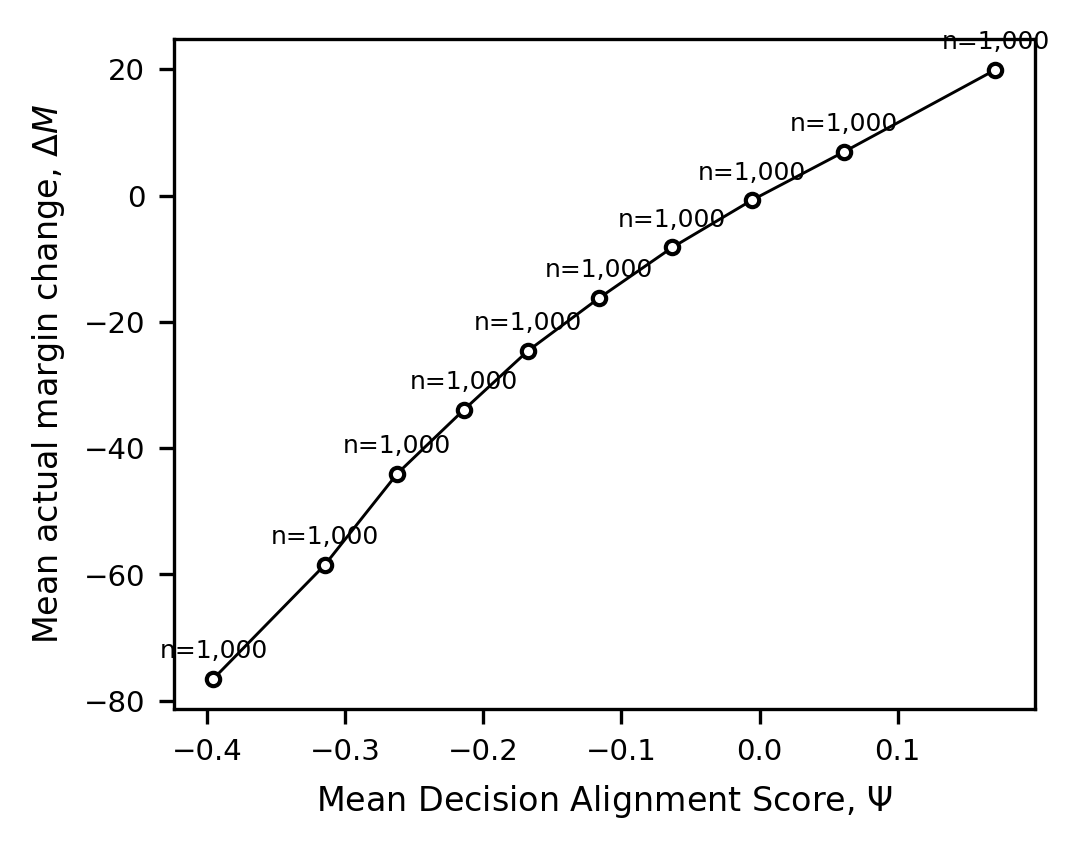}
\par\vspace{-1pt}
{\scriptsize S4}
\end{minipage}%
\end{minipage}\par

\vspace{1mm}
\caption{Equal-count DAS-group analysis for independently trained single-attention module models (S1--S4). Each point represents the mean DAS and actual margin change within a group; annotations indicate group sizes. Best viewed with zoom factor for clarity.}
\label{fig:cifar-deciles-single}

\end{figure*}

\section{Results and Discussion}
\label{sec:cifar-results-discussion}

\subsection{Recognition Performance}

Tables~\ref{tab:all-recognition} and \ref{tab:stage-recognition} summarize the classification performance of the ALL-trained models across five CNN backbones and the independently trained single-attention module configurations, respectively. In the ALL configuration, both SimAM and CBAM improve Top-1 accuracy over the attention-free baseline (B0) on four of the five backbones. The largest gain is obtained by ResNet-50 with SimAM, increasing Top-1 accuracy from 75.96\% to 79.11\%. These results show that the recognition benefit of attention depends on both the attention mechanism and the backbone.
The single-attention module results further reveal the influence of attention placement strategy. SimAM at S1 achieves the highest Top-1 accuracy of 77.99\%, compared with 76.88\% for B0, while the remaining placements yield smaller accuracy changes. Thus, the recognition effect of attention varies across network stages. However, classification accuracy alone does not characterize how attention-induced feature changes affect downstream decisions. We thus examine recognition performance separately from margin prediction agreement and decision-space alignment.

\subsection{Local Prediction and Margin Response}

Table~\ref{tab:decision-analysis} and Figs.~\ref{fig:cifar-prediction-all} and \ref{fig:cifar-prediction-single} assess the agreement between predicted and actual margin changes. In the ALL-last configurations, most backbone--attention combinations exhibit good predicted--actual correlations and relatively low prediction errors. For CBAM, this behavior is consistent across the four ResNet backbones, whereas MobileNetV2 exhibits comparatively larger errors. SimAM also shows the agreement across the evaluated backbones, although the prediction quality varies with architecture. The scatter plots in Fig.~\ref{fig:cifar-prediction-all} provide a complementary view of these relationships. Overall, the correlation and error measures indicate that the local first-order approximation captures the actual margin response reasonably well in most ALL-last configurations.

The single-attention module experiments reveal a dependence on attention placement (Fig.~\ref{fig:cifar-prediction-single}). For SimAM, S1 exhibits a negative predicted--actual correlation ($r=-0.16$), while S2 shows only a weak positive correlation ($r=0.28$). CBAM at S1 also exhibits a negligible negative correlation ($r=-0.04$). Although CBAM at S2 achieves a higher correlation ($r=0.68$), its prediction error remains relatively large (MAE = 14.39), indicating that correlation alone does not ensure numerical agreement.

For both attention mechanisms, higher predicted--actual correlations are observed at S3 and S4. These results suggest that the reliability of the local margin approximation depends on the location of the analyzed attention module, with later-stage configurations generally exhibiting better correspondence between predicted and actual margin changes.

\subsection{Directional Alignment and Activation Changes}

Table~\ref{tab:decision-analysis} and Figs.~\ref{fig:cifar-alignment-all}--\ref{fig:cifar-deciles-single} examine the association between DAS and actual margin change across attention mechanisms and network configurations. In the ALL-last experiments, both SimAM and CBAM exhibit positive correlations across the five CNN backbones. The DAS-versus-margin-change plots in Fig.~\ref{fig:cifar-alignment-all} show these sample-wise associations, while the equal-count DAS-group analysis in Fig.~\ref{fig:cifar-deciles-all} summarizes the corresponding trend across increasing levels of directional alignment. 

These results indicate that DAS is associated with the actual margin response in the evaluated ALL-last configurations.
The single-attention module experiments also reveal a dependence on attention placement. As shown in Fig.~\ref{fig:cifar-alignment-single}, DAS exhibits weak negative correlations at S1 for both attention mechanisms, while CBAM at S2 also shows a weak association with actual margin change. In contrast, higher positive correlations are observed at S3 and S4. Fig.~\ref{fig:cifar-deciles-single} further illustrates how the average margin response varies with DAS across attention locations. Downstream ReLU and ReLU6 activation states also alter under both attention mechanisms, indicating possible transitions across local affine regions.

\section{Conclusion}
\label{sec:conclusion}

This paper presents a decision-space framework for analyzing the effects of attention-induced feature perturbations on CNN classification margins. We derive a first-order margin prediction and introduce the Decision Alignment Score to quantify directional alignment with the local decision direction. Experiments with SimAM and CBAM across different CNN backbones show good prediction agreement and positive DAS correlations in most configurations. Single-attention module experiments reveal that both relationships depend on attention placement, with associations generally observed at later stages.
Margin prediction and directional alignment provide complementary measures for analyzing the effects of visual attention beyond recognition accuracy and visualization maps. However, the evaluation is limited to CIFAR-100, two attention mechanisms (SimAM and CBAM), and five CNN backbones. Future work will extend the framework to larger-scale datasets, diverse attention mechanisms, and modern architectures. These directions may contribute to a broader understanding of how visual attention influences deep representations and downstream decisions.

\bibliographystyle{IEEEtran}
\bibliography{references}

\vspace{2em}
\noindent
\begin{minipage}[t]{0.22\columnwidth}
    \vspace{0pt}
    \includegraphics[
        width=\linewidth,
        height=2.7cm,
        keepaspectratio
    ]{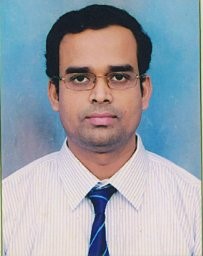}
\end{minipage}%
\hfill
\begin{minipage}[t]{0.74\columnwidth}
    \vspace{0pt}
    \normalfont\small
    \textbf{Niladri B. Puhan} received the B.E. degree in Electrical
    Engineering from UCE Burla, India, in 2000, the M.E. degree in
    Signal Processing from the Indian Institute of Science,
    Bangalore, India, in 2002, and the Ph.D. degree from Nanyang
    Technological University, Singapore, in 2007. He is currently an
    Associate Professor with the School of Electrical and Computer
    Sciences, Indian Institute of Technology Bhubaneswar, India.
    His research interests include signal and image processing,
    biometrics, medical image analysis, and deep learning.
\end{minipage}

\vspace{2em}

\noindent
\begin{minipage}[t]{0.22\columnwidth}
    \vspace{0pt}
    \includegraphics[
        width=\linewidth,
        height=2.7cm,
        keepaspectratio
    ]{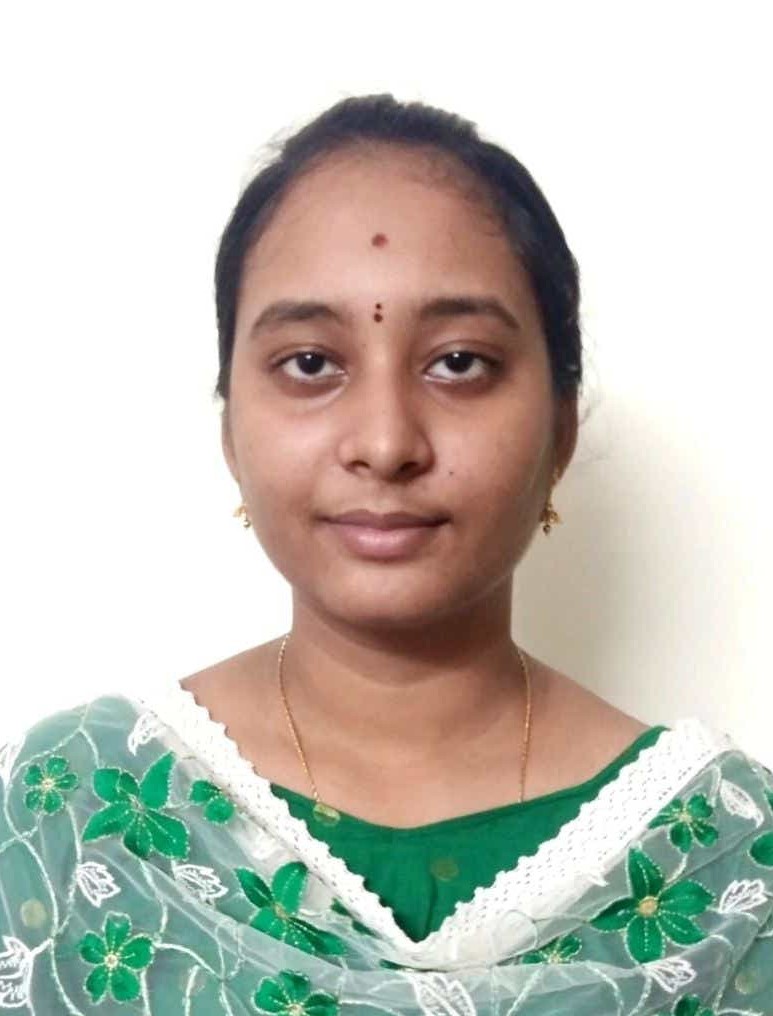}
\end{minipage}%
\hfill
\begin{minipage}[t]{0.74\columnwidth}
    \vspace{0pt}
    \normalfont\small
    \textbf{K. Kamala Priya} received the B.Tech. degree in
    Electronics and Communication Engineering from JNTU
    Kakinada, India, in 2022. She is currently a research
    scholar in the School of Electrical and Computer Sciences,
    IIT Bhubaneswar, India. Her research interests include
    signal and image processing, medical image analysis,
    and deep learning.
\end{minipage}

\end{document}